\documentclass[10pt,twocolumn,letterpaper]{article}

\usepackage{wacv}              

\usepackage{graphicx}
\usepackage{amsmath}
\usepackage{amssymb}
\usepackage{booktabs}
\usepackage{times}
\usepackage{epsfig}
\usepackage[table,x11names,dvipsnames]{xcolor}
\usepackage{svg}
\usepackage{macros}
\usepackage{siunitx}
\usepackage{multirow}
\usepackage[accsupp]{axessibility}
\usepackage[pagebackref,breaklinks,colorlinks]{hyperref}

\usepackage[capitalize]{cleveref}
\crefname{section}{Sec.}{Secs.}
\Crefname{section}{Section}{Sections}
\Crefname{table}{Table}{Tables}
\crefname{table}{Tab.}{Tabs.}

\def\wacvPaperID{1807} 
\def\confName{WACV}
\def\confYear{2024}

\begin{document}

\title{Re-Evaluating LiDAR Scene Flow}

\author{Nathaniel Chodosh\\
Carnegie Mellon University\\
{\tt\small nchodosh@andrew.cmu.edu}
\and
Deva Ramanan\thanks{Equal contribution.}\\
Carnegie Mellon University\\
{\tt\small deva@cs.cmu.edu}
\and
Simon Lucey$^*$\\
University of Adelaide\\
{\tt\small simon.lucey@adelaide.edu.au}
}
\maketitle

\begin{abstract}
    Popular benchmarks for self-supervised LiDAR scene flow (stereoKITTI, and FlyingThings3D) have unrealistic rates of dynamic motion, unrealistic correspondences, and unrealistic sampling patterns. As a result, progress on these benchmarks is misleading and may cause researchers to focus on the wrong problems.
    We evaluate a suite of top methods 
    on a suite of real-world datasets (Argoverse 2.0, Waymo, and NuScenes)
    and report several conclusions. First, we find that performance on stereoKITTI is negatively correlated with performance on real-world data.
 Second, we find that one of this task's key components -- removing the dominant ego-motion -- is better solved by classic ICP than any tested method. Finally, we show that despite the emphasis placed on learning, most performance gains are caused by pre- and post-processing steps: piecewise-rigid refinement and ground removal. We demonstrate this through a baseline method that combines these processing steps with a learning-free test-time flow optimization. This baseline outperforms every evaluated method. 
\end{abstract}
\begin{figure}
    \centering
    \includegraphics[width=0.49\textwidth,trim= 0cm 0cm 0cm 0cm,clip]{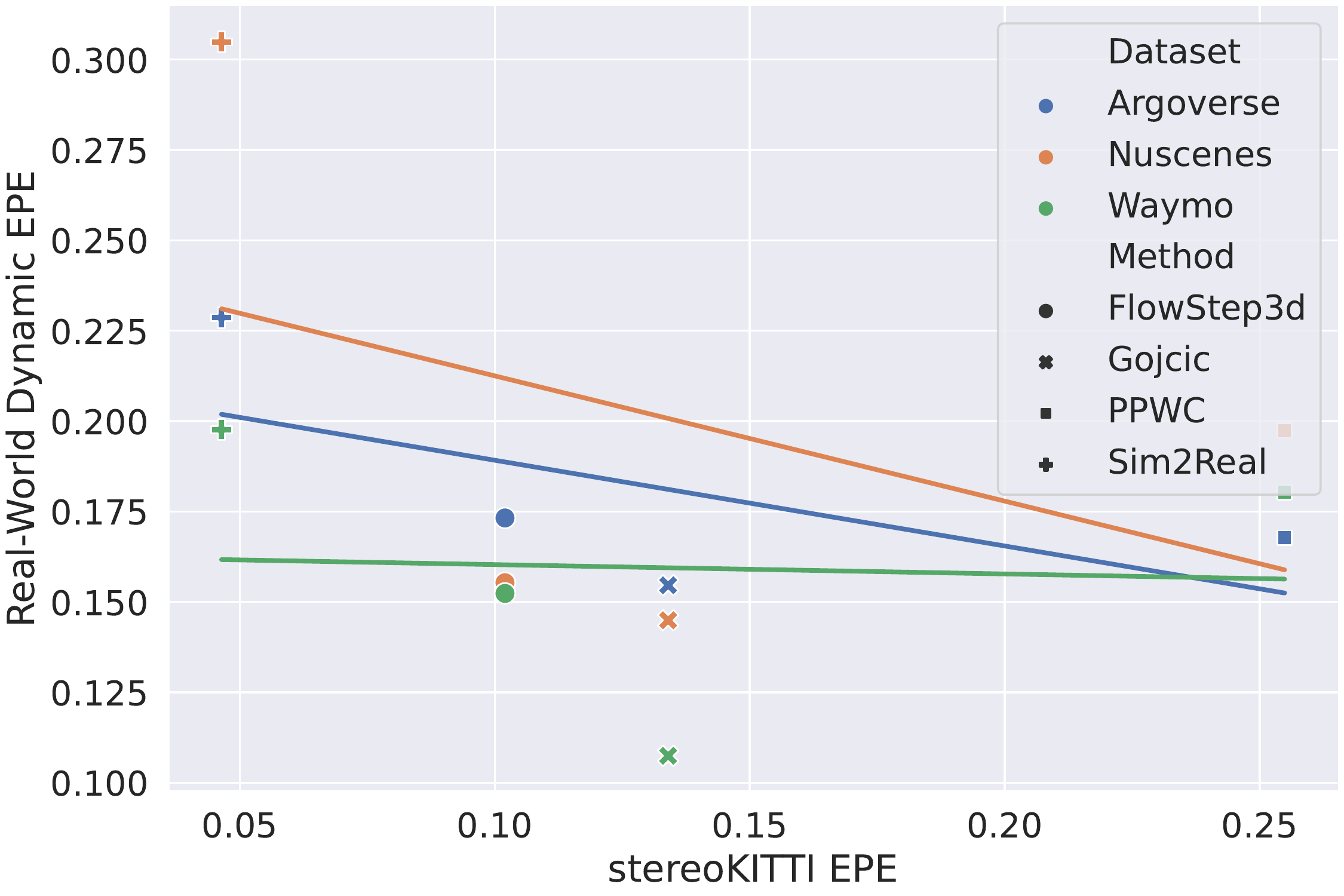}
    \caption{Recent self-supervised scene flow methods\cite{wu2020pointpwc,kittenplon2021flowstep3d,jin2022deformation,gojcic2021weakly} typically 
     focus on performance on stereoKITTI\cite{liu2019flownet3d}. We call attention to several problematic aspects of its semi-synthetic construction. When evaluated on real-world datasets (Waymo, Argoverse, NuScenes) 
     we find a \textbf{negative correlation with stereoKITTI performance}. For each method and LiDAR dataset, we plot a point corresponding to the self-reported end-point error on stereoKITTI versus the end-point error on dynamic points of that method trained on real data. For each dataset, we plot the best-fit line to visualize the correlation.}
    \label{fig:perf-over-time}
\end{figure}
\section{Introduction}
\label{sec:intro}
Research is often guided by improving benchmark performance, but we show that popular scene flow benchmarks may be guiding research in the wrong direction. Scene flow, the 3D analog of optical flow~\cite{vedula1999three}, can help autonomous vehicles identify moving objects. Most autonomous vehicles use LiDAR sensors for 3D perception, creating interest in estimating the dynamic motion between successive scans~\cite{liu2019flownet3d,behl2019pointflownet,baur2021slim,gojcic2021weakly,Mittal_2020_CVPR,jund2021scalable}. If this task can be accomplished without relying on labeled data, it can give autonomous vehicles awareness of moving objects outside the detection taxonomy\cite{najibi2022motion}. The potential for label-free detection has led to significant interest in self-supervised scene flow methods~\cite{wu2020pointpwc,Mittal_2020_CVPR,baur2021slim,jin2022deformation}. We focus on this self-supervised setting and show that the standard benchmarks have several fundamental flaws. When evaluated more realistically, apparent benchmark improvements correspond to stalled or even decreasing real-world performance (\cref{fig:perf-over-time}).

Most work follows the evaluation framework from FlowNet3D~\cite{liu2019flownet3d}, which primarily measures the average end-point-error (EPE) across FlyingThings3D~\cite{mayer2016large} and KITTI-SF~\cite{menze2015joint} (now typically referred to as stereoKITTI). These datasets have several issues:
(1) stereoKITTI samples dynamic objects with an artificial pattern that differs from the pattern on static objects, inadvertently providing part of the ``answer" to learning-based approaches.
(2) Both FlyingThings3D and stereoKITTI ensure successive point clouds are in one-to-one correspondence, which is never the case for actual scans. 
(3) Both datasets contain an unrealistically high percentage of dynamic points. In contrast, real-world data is dominated by the background.

Together these issues obfuscate the main challenges of LiDAR scene flow: identifying the few non-static points and estimating their motions robustly given the lack of correspondences. We demonstrate the impact of these issues by evaluating a suite of top methods on several large-scale real-world datasets (Argoverse, NuScenes, Waymo), finding that performance is \emph{negatively correlated} with performance on the standard benchmarks.

We also believe these benchmarks have led researchers to ignore the virtues of classic optimization-based approaches. Many learning-based methods have been proposed for first estimating and removing ego-motion\cite{tishchenko2020self,gojcic2021weakly}, but we show that none outperform Iterative Closest Point (ICP). Furthermore, we show that the common pre- and post-processing steps of ground removal and enforcing piecewise-rigidity have a larger impact on performance than any learning strategy. We show this through a baseline method that combines these steps with a test-time optimization flow method. This baseline, without any learning, outperforms every method in our suite and outperforms all self-supervised methods on the self-reported 
NuScenes~\cite{mayer2016large} and lidarKITTI~\cite{gojcic2021weakly} benchmarks. This leads to the conclusion that current learning methods fail to extract information not present in a single example despite using large amounts of data.
 
In summary, our main contributions are:
\begin{itemize}
\item An investigation of the weakness of current self-supervised scene flow evaluations.
\item A \emph{dataless} flow method which gives state-of-the-art results.
\item A standard evaluation protocol and codebase along with a ``model-zoo'' of top methods as well as flow labels for Argoverse 2.0\footnote{Code, weights, and outputs for all the evaluated methods will be released}.

\end{itemize}

\section{Related Work}
\textbf{Scene Flow:} Scene flow was introduced by \cite{vedula1999}, who posed the problem in the stereo RGB setting and spawned a large body of subsequent work~\cite{basha2013multi,chen1992object,hadfield2011kinecting,hornacek2014sphereflow,hadfield2013scene,menze2015object,pons2003variational,pons2007multi,vogel20113d,vogel2013piecewise}. A related problem is non-rigid registration~\cite{amberg2007optimal,chui2003new,li2008global,izadi2011kinectfusion,pauly2005example}, which is focused on fitting dense point clouds or meshes. We are interested in the setting without images, based purely on sparse LiDAR point clouds.

\textbf{Optimization based LiDAR Scene Flow:} Dataless LiDAR scene flow estimation was first proposed by Dewan \etal~\cite{dewan2016rigid}. Inspired by  a non-rigid registration method regularized by the graph Laplacian~\cite{eisenberger2020smooth},  Pontes \etal~\cite{pontes2020scene} created an improved method. Their results were further improved upon by Li \etal~\cite{li2021neural} with the implicit regularization of coordinate networks~\cite{atzmon2020sal,chen2019learning,mescheder2019occupancy,park2019deepsdf,sitzmann2020implicit}. We use ~\cite{li2021neural} as the backbone of our baseline and additionally employ coordinate networks for height-map estimation. Apart from these, the vast majority of recent methods have focused on deep learning based solutions.

\textbf{Self and Weakly-Supervised LiDAR Scene Flow:} ``Just Go with the Flow''~\cite{Mittal_2020_CVPR} demonstrated that using a combination of nearest-neighbor and cycle consistency losses was enough to train the FlowNet3D network, avoiding the reliance on labeled data. This led to many other works which adopted similar losses~\cite{wu2020pointpwc,tishchenko2020self,kittenplon2021flowstep3d,baur2021slim}. Others made use of easier to acquire sources of supervision such as foreground/background segmentation masks~\cite{gojcic2021weakly,dong2022exploiting} or addressed the synthetic to real domain gap~\cite{jin2022deformation}. Of particular relevance are those methods which make use of ego-motion estimation and piecewise rigid representations~\cite{gojcic2021weakly,dong2022exploiting,li2022rigidflow}. We show that these steps are critical to good performance, but find that ICP vastly outperforms learned approaches and that piecewise rigidity is more effective as a post-processing step than as a loss regularizer.

\textbf{Ground Segmentation:} In robotics, ground segmentation has been studied as a subset of general dataless segmentation~\cite{moosmann2009segmentation}, traversable area identification~\cite{himmelsbach2010fast}, and as a pre-processing step for object detection, classification and tracking~\cite{zermas2017fast,narksri2018slope,jimenez2021ground,lee2022patchwork++}. Current scene flow methods do not make use of these sophisticated methods and instead rely on basic plane fitting~\cite{baur2021slim,gu2019hplflownet,najibi2022motion}. In order to show how even small improvements to pre- and post- processing steps can drastically improve scene flow estimation, we propose a simple segmentation method based on coordinate networks.

\section{Benchmark Issues}
\label{sec:dataset-problems}
Most self-supervised flow estimation works inherit their evaluation protocol from \cite{liu2019flownet3d}, which is based on the synthetic FlyingThings3D~\cite{mayer2016large} and stereoKITTI~\cite{menze2015joint,geiger2012we}. Both were created for evaluating RGB-based flow methods and \cite{liu2019flownet3d} extended them to point clouds by lifting the optical flow and depth annotations to 3D. These datasets and protocols suffer from three main deficiencies.
\begin{figure}
    \centering
    \includegraphics[width=0.45\textwidth,trim=0 0 0 0,clip]{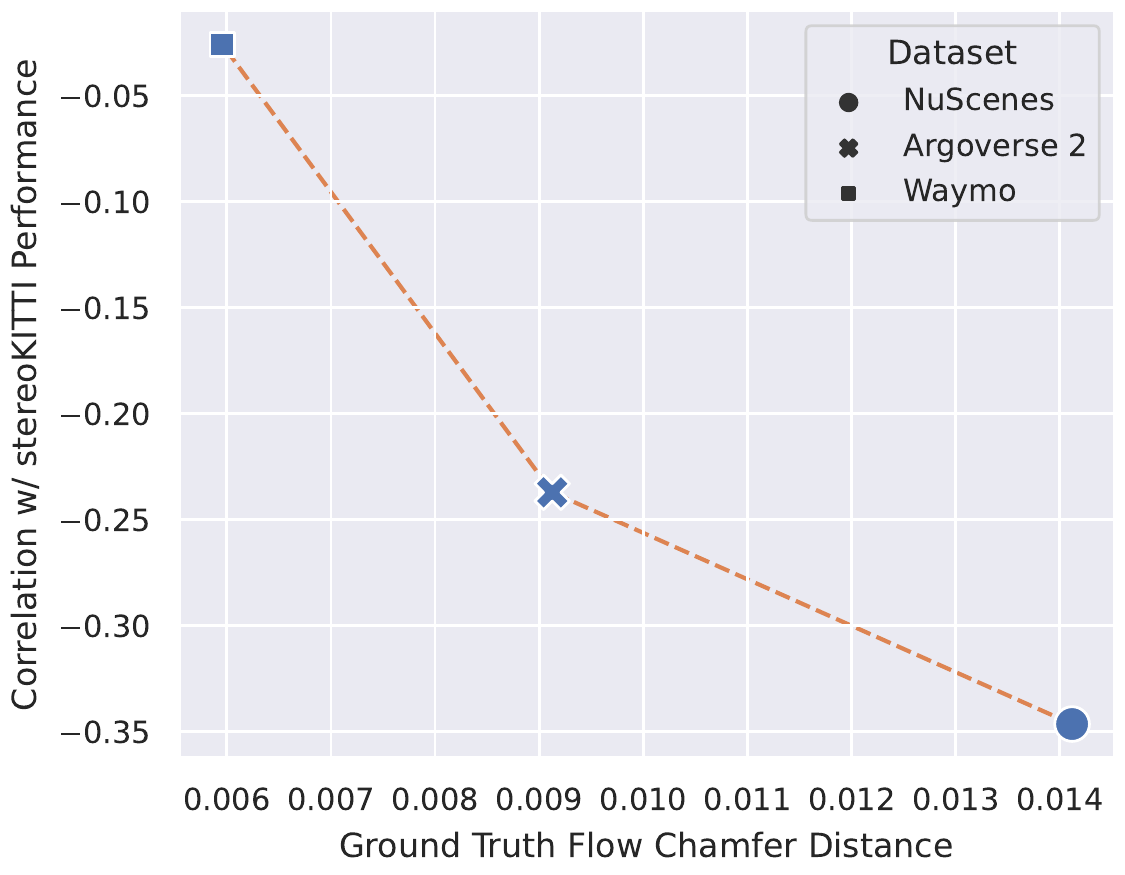}
    \caption{The correlation between performance on stereoKITTI and performance on real datasets as a function of how much those datasets violate the one-to-one correspondence assumption present in stereoKITTI. The more a dataset violates this assumption (higher chamfer distance due to sparser pointclouds), the worse the correlation with stereoKITTI performance.} 
    \label{fig:corr}
\end{figure}

\textbf{The one-to-one correspondence assumption removes the main challenge of working with LiDAR data.}
Since both FlyingThings3D and stereoKITTI are created by lifting optical flow annotations, the point clouds for each input pair are in one-to-one correspondence. For each 3D point $\p_i$ and ground truth 3D flow $\f_i$ in the first frame, there exists a 3D point $\q_i$ in the second frame such that $\p_i + \f_i = \q_i$. Real-world LiDAR scans do not have this property. The presence of correspondences fundamentally changes the LiDAR scene flow problem into a point-matching problem. 
It has been claimed that randomly sub-sampling the input breaks this correspondence. However, given the total number of points in each scan (90k) and the number of sub-sampled points (8192), there are still an expected 745 corresponding points in the input. Since each example has only a handful of independent motions, these correspondences are enough to constrain the solution.

This problem is especially severe for self-supervised scene flow methods trained using the chamfer distance, a symmetric extension of the nearest neighbor distance\cite{Mittal_2020_CVPR}. Essentially, each method deforms the first point cloud using the predicted flows and then computes the distance to the second point cloud. The chamfer distance is an excellent self-supervised objective for point clouds with one-to-one correspondences since the ground truth flow achieves the minimum distance of 0. However, for real-world point clouds with no correspondences the chamfer distance becomes a weaker proxy, and even the ground truth flow will have a non-zero distance. We can quantify this effect by computing the chamfer distance of the ground truth flows for several real-world datasets. In \cref{fig:corr} we show the relationship between that quantity and the slopes of the best-fit lines shown in \cref{fig:perf-over-time}. As can be seen, the more a dataset violates the one-to-one assumption, the more \emph{good} performance on stereoKITTI indicates \emph{poor} performance on that dataset.
\begin{figure}
    \centering
    \includegraphics[width=0.45\textwidth,clip]{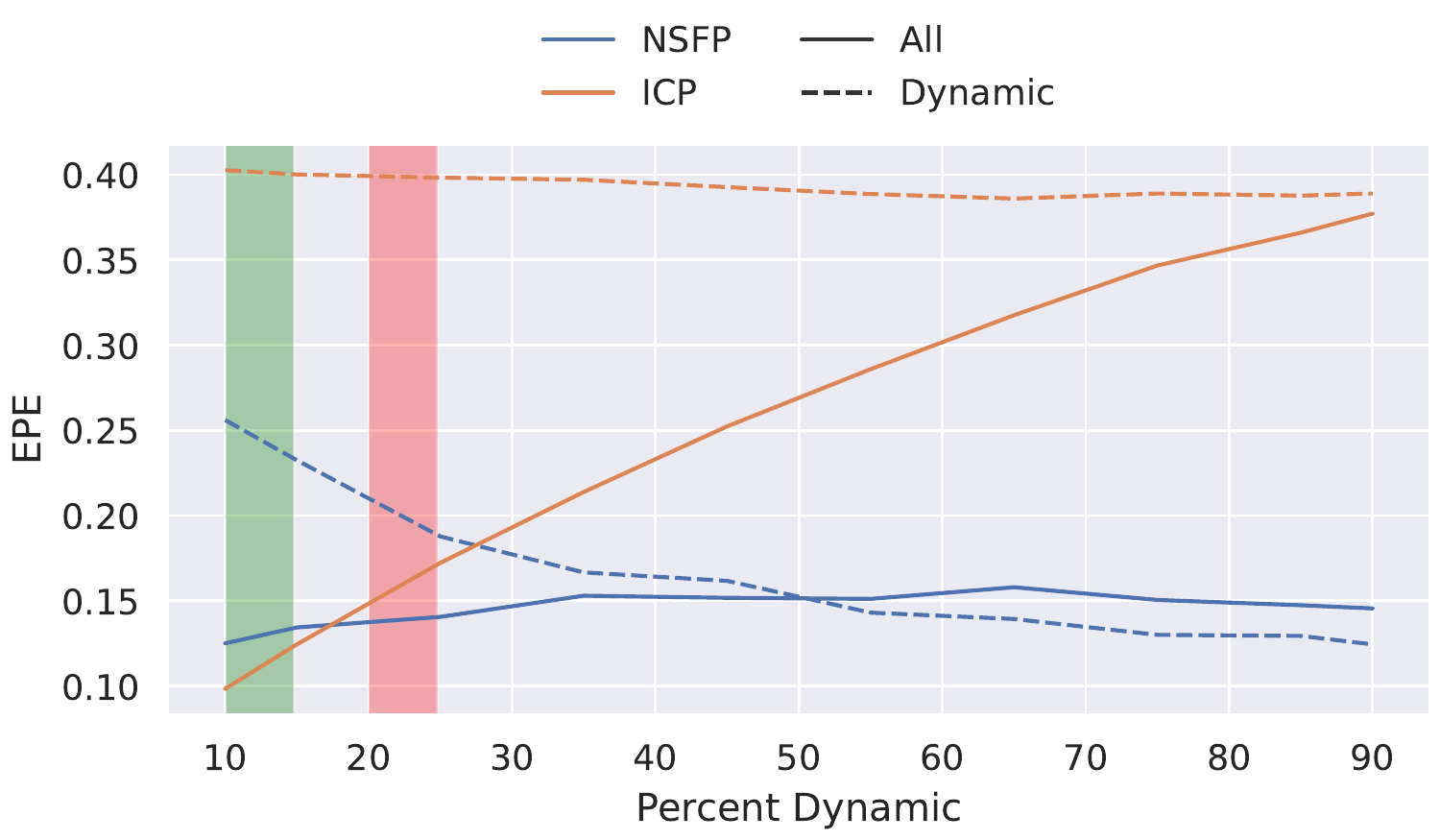}
    \caption{The performance of ICP and NSFP versus the ratio of dynamic points in each example. The green region indicates the ratio found in real data and the red region indicates the ratio found in the stereoKITTI dataset. The unrealistic dynamic ratio of stereoKITTI causes ICP to appear to perform worse than it does on real-world data such as Argoverse.} 
    \vspace{-0.35cm}
    \label{fig:dynamic}
\end{figure}
\begin{figure}
    \centering
    \includegraphics[width=\linewidth,clip]{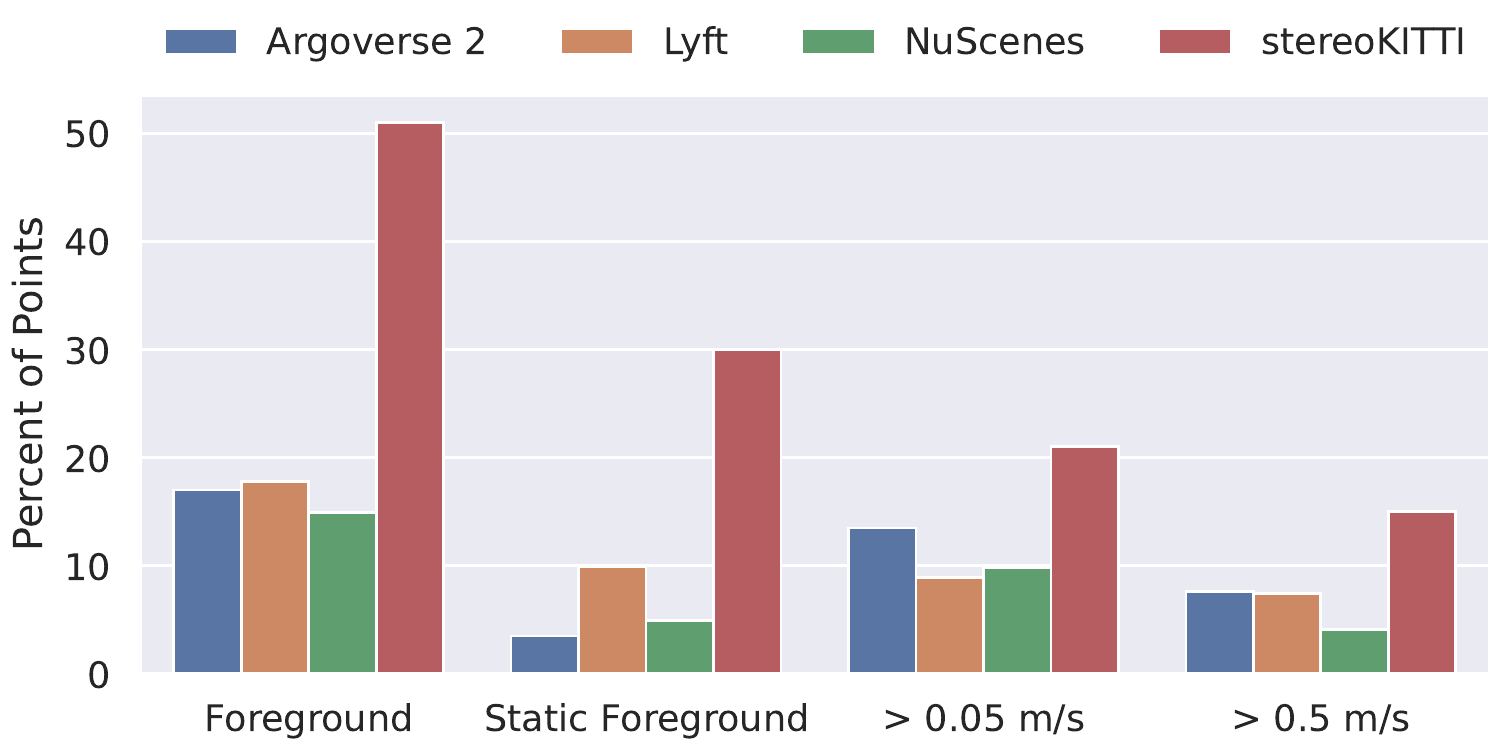}
    \caption{An analysis of the motion profile of points found in various autonomous driving datasets. The leftmost column is the percentage of points belonging to any tracked object. The right three columns show those points separated by their speed once ego-motion has been removed. All three datasets have a very low rate of dynamic foreground points, in contrast to the typical stereoKITTI benchmark.}
    \label{fig:datasets} 
\end{figure}

\textbf{Current benchmarks' unrealistic rates of dynamic motion make estimating ego-motion seem harder than it is.} 
In any scene, some portion of the measured points belong to the static background rather than dynamically moving objects. In real-world LiDAR scans such as those in NuScenes~\cite{caesar2020nuscenes}, Waymo~\cite{jund2021scalable}, or Argoverse 2~\cite{wilson2021argoverse}, the percentage of static points is very high, ranging from 85-100\% (\cref{fig:datasets}). In contrast, FlyingThings3D has 0\% static points since it consists of flying things, and the KITTI-SF dataset has approximately 75-85\% static points. Dealing with this data imbalance is one of the key difficulties of self-supervised flow estimation but it is not present in the popular benchmarks. This discrepancy also explains why ICP has been ignored as a technique for ego-motion estimation\cite{liu2019flownet3d}.

In order to understand why recent methods have neglected ICP, we need to understand how the amount of dynamic points in a scene affects the performance of ICP. We manually re-sampled each example in the Argoverse 2.0 validation dataset to take a specified number of dynamic and static points and then ran a test-time optimisation scene flow method~\cite{li2021neural} and ICP~\cite{chen1992object}. The results (\cref{fig:dynamic}) show that when looking at the total EPE averaged over the entire dataset, like most evaluations, the performance of ICP steadily degrades as the percentage of dynamic points increases. At 20\% dynamic points, double the rate of real-world data but approximately the same ratio as KITTI-SF, ICP's performance has degraded to significantly below the performance of NSFP. This explains why early methods which did include ICP in their comparisons~\cite{liu2019flownet3d} were able to outperform it, leading future methods to leave it out even as more realistic data became available.
\begin{figure}
    \centering
    \includegraphics[width=0.48\linewidth]{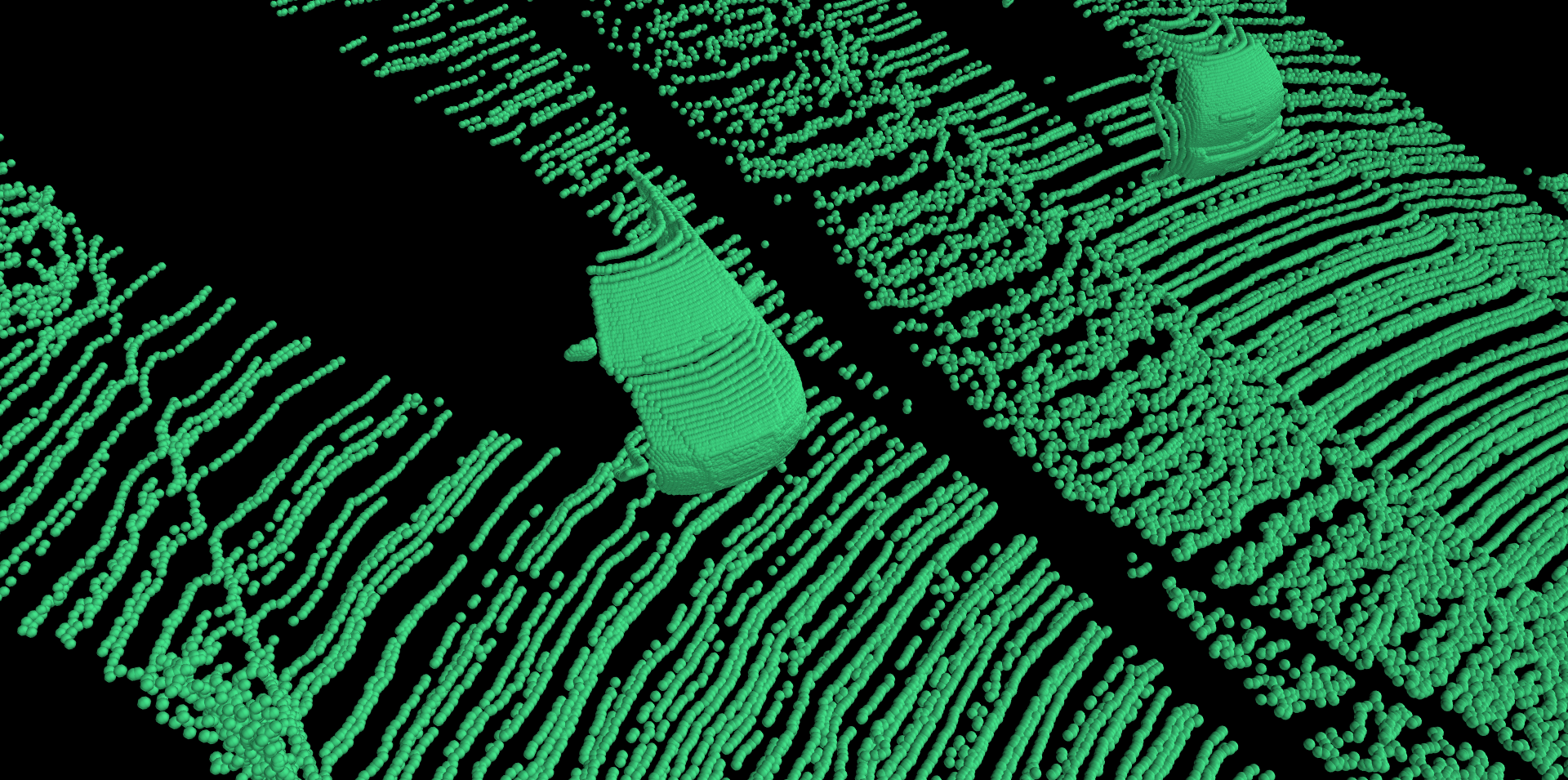}
    \includegraphics[width=0.48\linewidth]{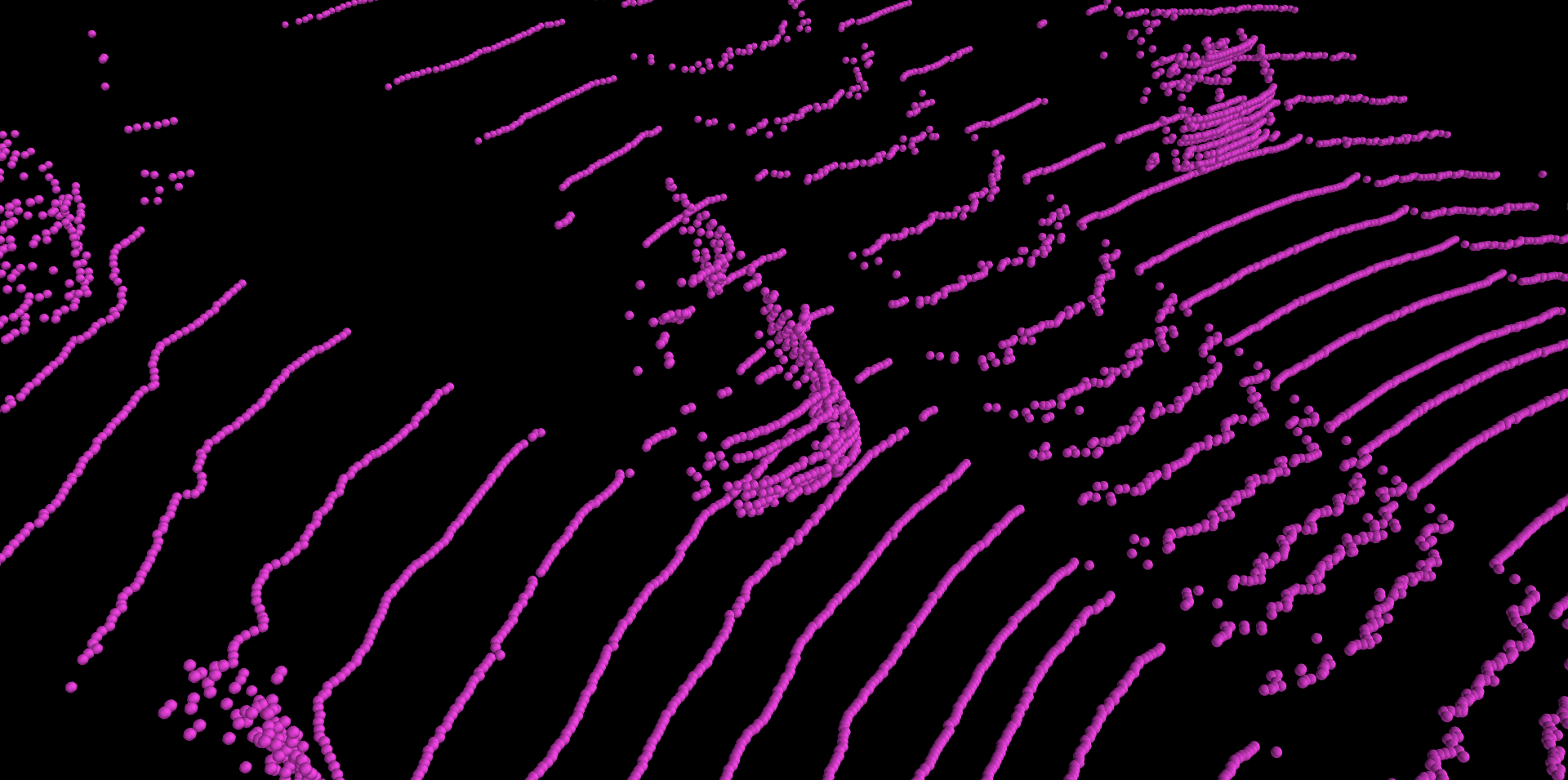}
    \caption{Comparison of sampling patterns from KITTI-SF (\textbf{left}) versus the corresponding real LiDAR scan (\textbf{right}). KITTI-SF uses dense CAD models for foreground objects, which makes it easier for learning-based methods to find them among the sparse LiDAR background points.}
    \label{fig:kitti}
\end{figure}
\begin{figure}
    \centering
    \includegraphics[width=0.23\textwidth, trim=20cm 10cm 20cm 10cm,clip]{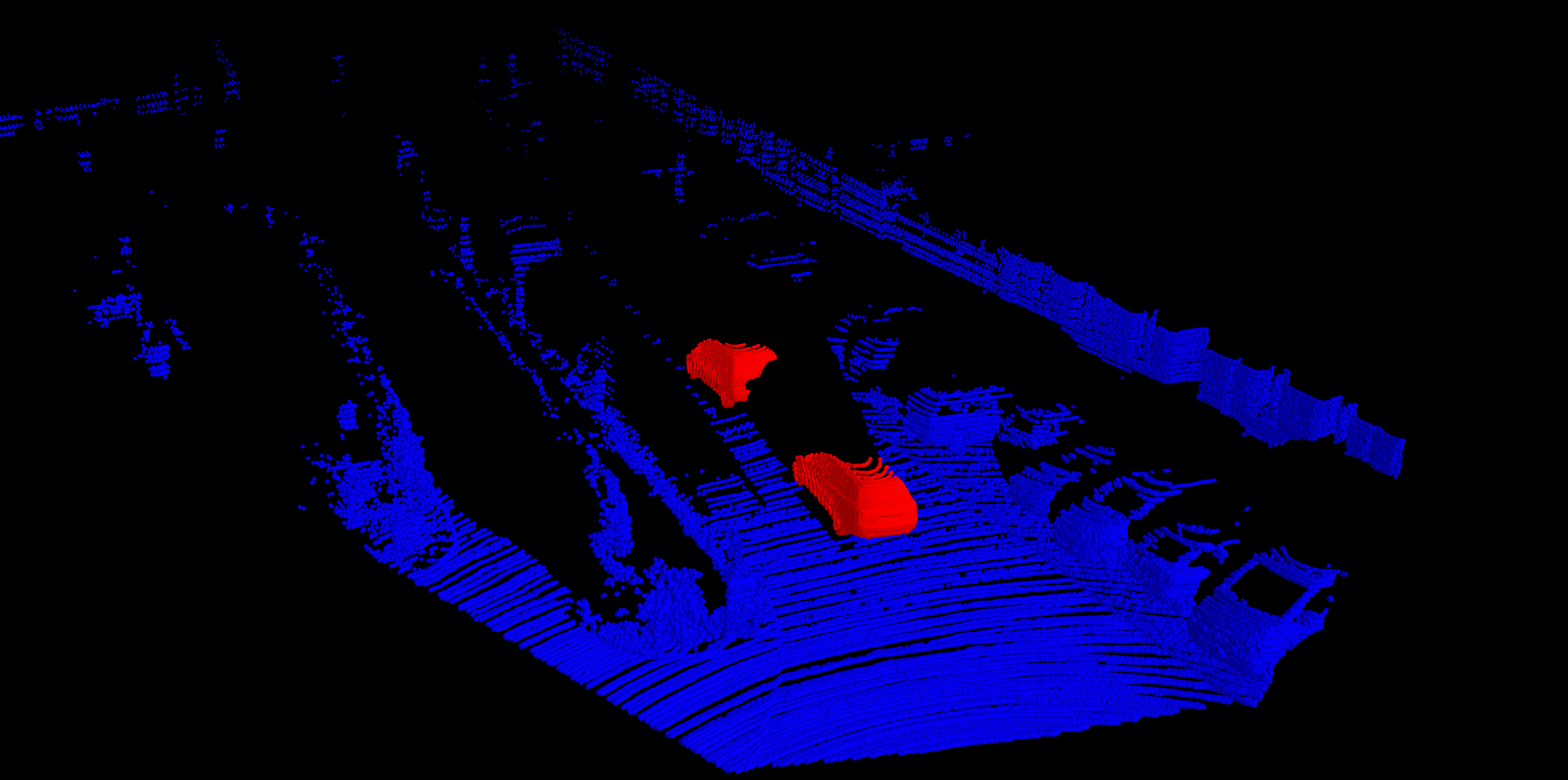}
    \includegraphics[width=0.23\textwidth, trim=20cm 10cm 20cm 10cm,clip]{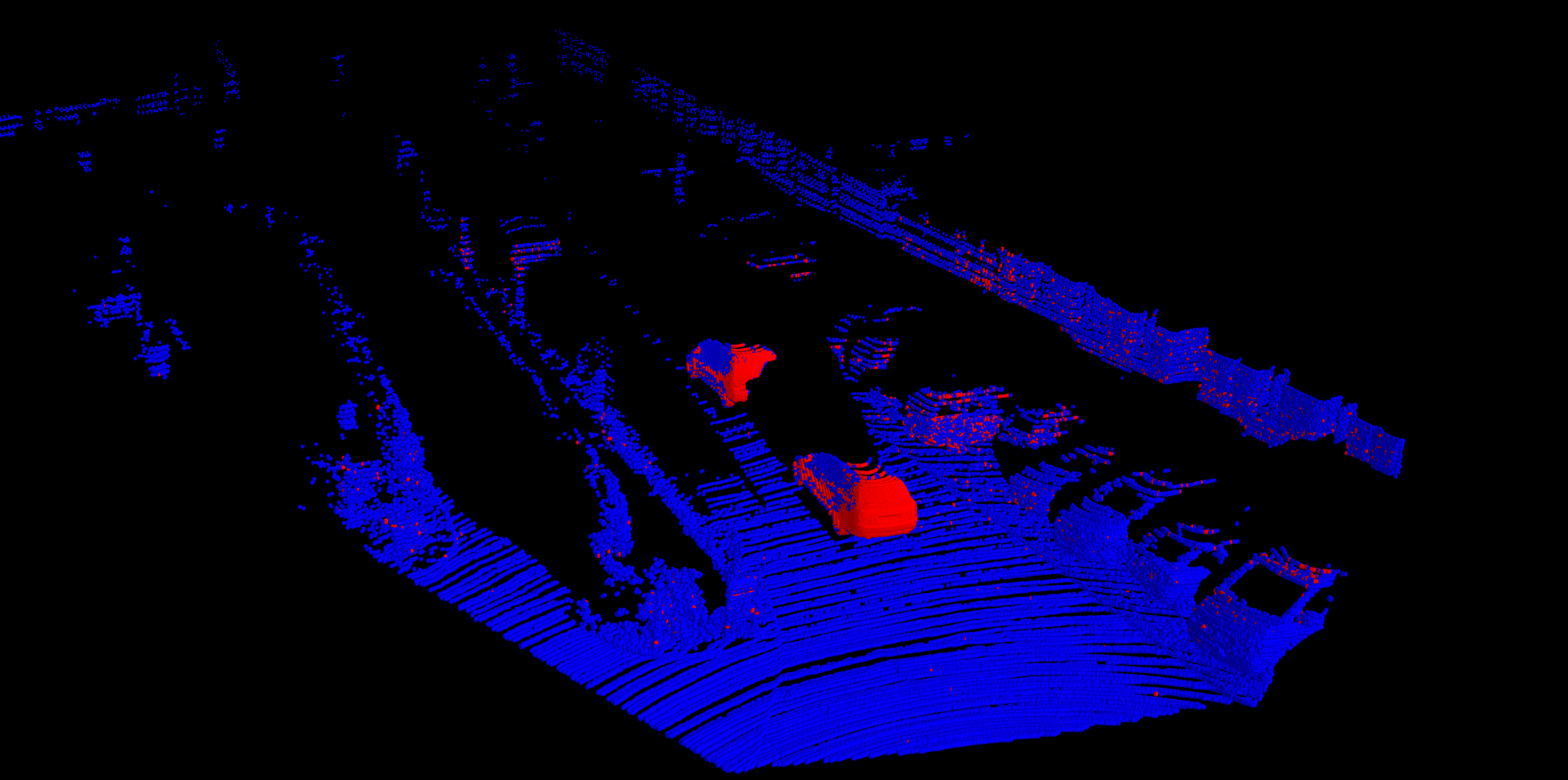}
    \caption{An example ground truth segmentation from KITTI-SF (\textbf{left}) compared with (\textbf{right}) an example prediction from our local model. Note how the model can even identify parked vs moving cars since only moving vehicles have the dense sampling pattern.}
    \label{fig:kitti-prediction}
    \vspace{-0.2cm}
\end{figure}

\textbf{The sampling pattern of dynamic objects in stereoKITTI gives away the answer in the input.} 
The ground truth flow and disparity of KITTI-SF were created by fitting 3D models to the LiDAR points and 2D annotations~\cite{menze2015object}. This gives the dynamic objects a distinctive dense sampling pattern that separates them from the background (see \cref{fig:kitti}). However, determining which points belong to moving objects is one of the main challenges and the sampling pattern effectively gives away the answer in the input. We demonstrate that learning-based system can simply identify moving objects by the sampling pattern without learning anything about the motion.

\textbf{Model:} We use a model that takes a small amount of local context and \emph{no information from the second frame}, to predict motion segmentation for a point $\p$. That is, the input to our network is the set of vectors $\{\q - \p \mid \forall \q \in \mathcal{N}_r(\p)\}$ where $\mathcal{N}_r$ is a ball of radius $r$ centered at $\p$. We use $r = \qty{0.05}{\metre}$ in our experiment. Then we use a standard PointNet architecture to predict a binary label for the point.

\textbf{Training:} We train our model on the first 150 examples of KITTI-SF and test on the remaining 50. Note this is essentially the split used by FlowNet3D\cite{liu2019flownet3d} and the recent RigidFlow\cite{li2022rigidflow} for fine-tuning. We train our network using a cross-entropy loss weighted by the inverse frequency of each label. For comparison, we perform the same experiment on real LiDAR scans from Argoverse. We sample 150 scans from the train set and 50 scans from the validation set.

\textbf{Results:} Using only information in a \qty{5}{cm} ball around each point we are able to segment KITTI-SF with high accuracy. We achieve a mean intersection over union on the foreground of 0.83 and 0.97 on the foreground. An example is visualized in \cref{fig:kitti-prediction}. In comparison when we attempt this on real LiDAR scans that do not have a biased sampling pattern, we achieve foreground and background mean intersection over union scores of 0.16 and 0.81 respectively. This demonstrates that the sampling pattern of KITTI-SF makes it trivial to identify foreground points and should not be used for training or validation.

\subsection{Real-World Flow Evaluation}

Due to these deficiencies, recent works have made progress on improving results on these benchmarks without improving results on real-world data (\cref{fig:perf-over-time}). Some of the problems could perhaps be addressed through re-sampling, but doing so requires modifying the data based on the ground truth static-vs-dynamic labels. This runs counter to the goal of creating methods that operate on raw, unlabeled data. 

Some works have evaluated on real LiDAR data by transferring synthetic labels~\cite{gojcic2021weakly}, or more commonly by computing flows from object-level tracks~\cite{baur2021slim,jund2021scalable,jin2022deformation,li2021neural}. However, those evaluations have been presented as auxiliary results~\cite{baur2021slim,li2021neural} with limited comparisons to existing methods, or without comparison entirely~\cite{jund2021scalable}. As a result, KITTI-SF and FlyingThing3D remain the standard benchmarks. We argue that evaluating on real-data should be the ``gold-standard'' for scene flow as opposed to synthetic benchmarks.

To further emphasise how top learning-based methods fail when evaluated properly on real data, we also demonstrate that they can be outperformed by a simple test-time optimization baseline. We construct this baseline by making small improvements to standard pre- and post-processing steps and wrapping them around neural scene flow prior.
 
\section{Baseline}
\begin{figure}
    \centering
    \includegraphics[width=\linewidth,trim=0 8.5cm 7cm 0cm,clip]{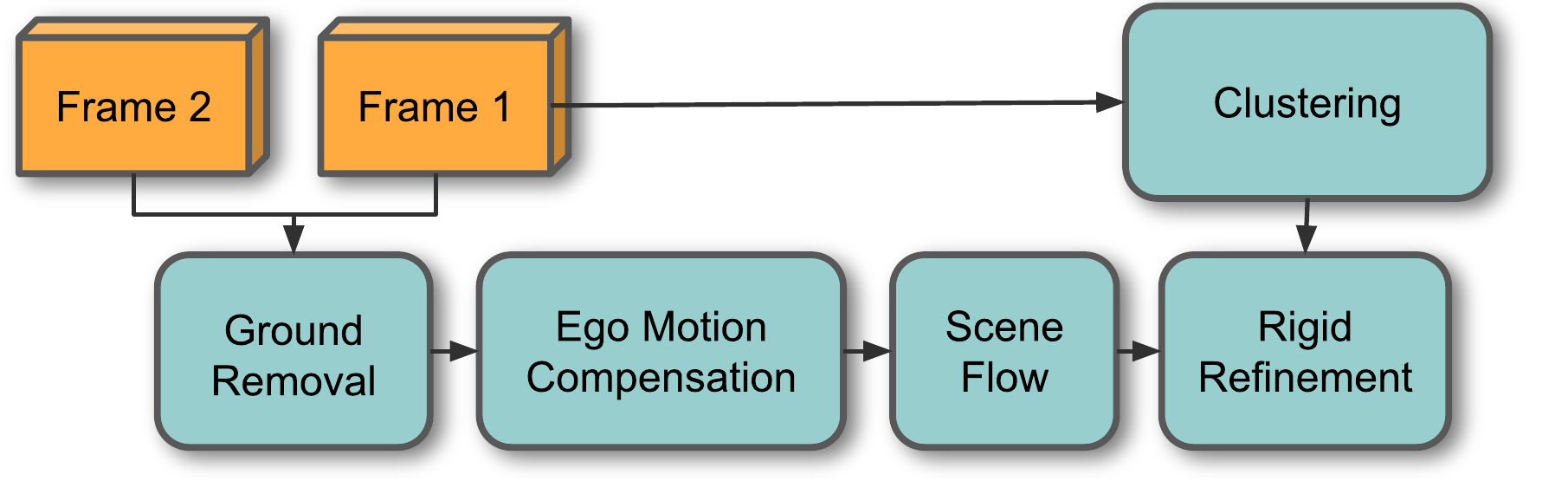}
    \caption{Our proposed pipeline for LiDAR scene flow based largely on classical pre- and post-processing.}
    \label{fig:pipeline}
\end{figure}
Our pipeline (\cref{fig:pipeline}), consists mainly of pre- and post-processing steps around the test-time flow optimization method of \cite{li2021neural}.
\begin{figure*}
    \centering
    \includegraphics[width=0.9\linewidth, trim=0 0.25cm 0cm 0,clip]{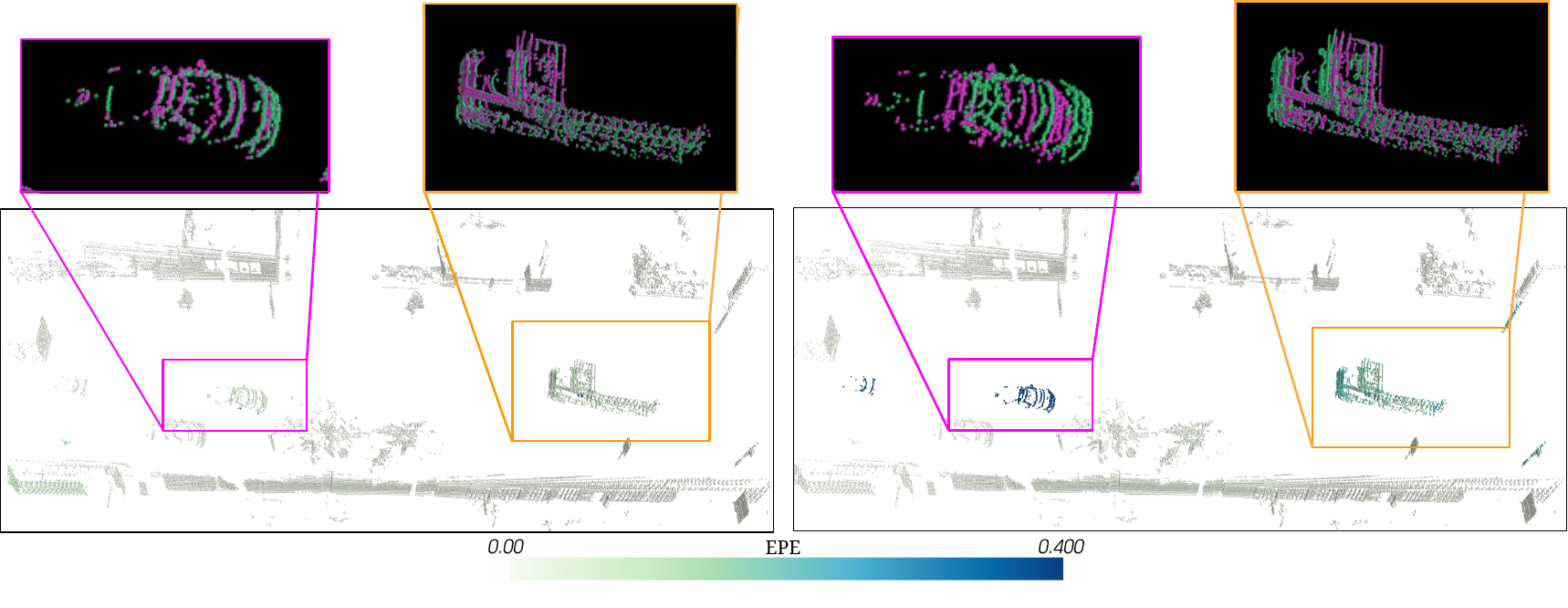}
    \caption{Comparison between NSFP with our proposed pre and post processing steps (\textbf{left}) and standard NSFP (\textbf{right}). In the (\textbf{bottom}) views points are color coded by EPE. The (\textbf{top}) detail views show the first and second frames aligned by the predicted flow. NSFP struggles to represent both foreground and background motion. We find that first using ICP to remove ego-motion greatly improves the estimates on dynamic points.}
    \label{fig:motion-comp}
\end{figure*}

\textbf{Motion Compensation:} Since real-world scenes consist mostly of static background objects, removing the ego-motion of the sensor makes estimating the dynamic motion significantly easier (\cref{fig:motion-comp}). For datasets such as Argoverse or Waymo, the ego-motion is provided and we transform the first scan into the coordinate frame of the second. NuScenes and lidarKITTI also have this information but previous work has opted to not use it. In this case, we use ICP~\cite{chen1992object,vizzo2023ral} to first estimate the ego-motion. We demonstrate experimentally that this is much more effective than learned approaches.
\begin{figure*}
    \centering
    \includegraphics[width=0.3\textwidth]{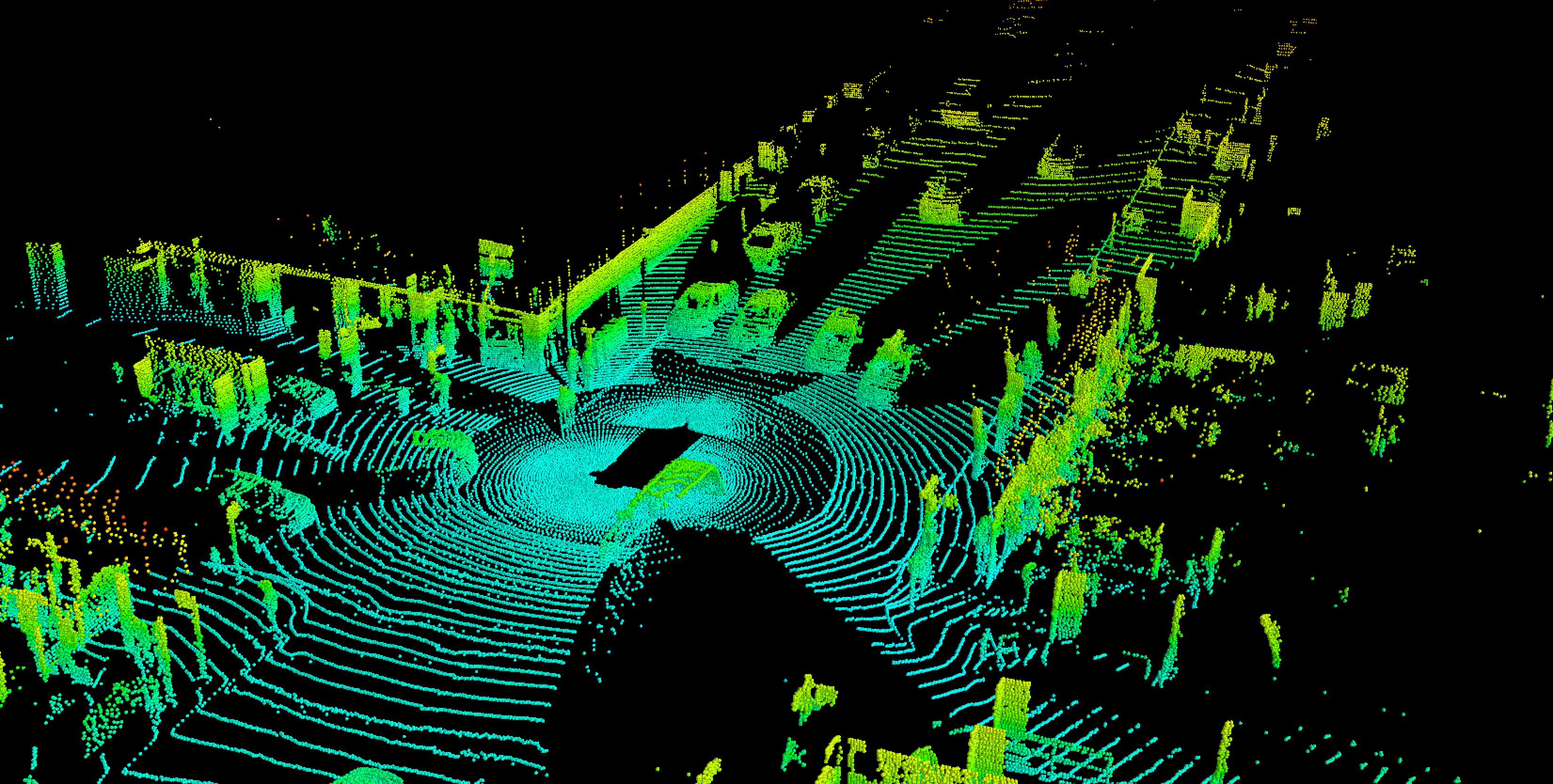}\hspace{0.05cm}
    \includegraphics[width=0.3\textwidth]{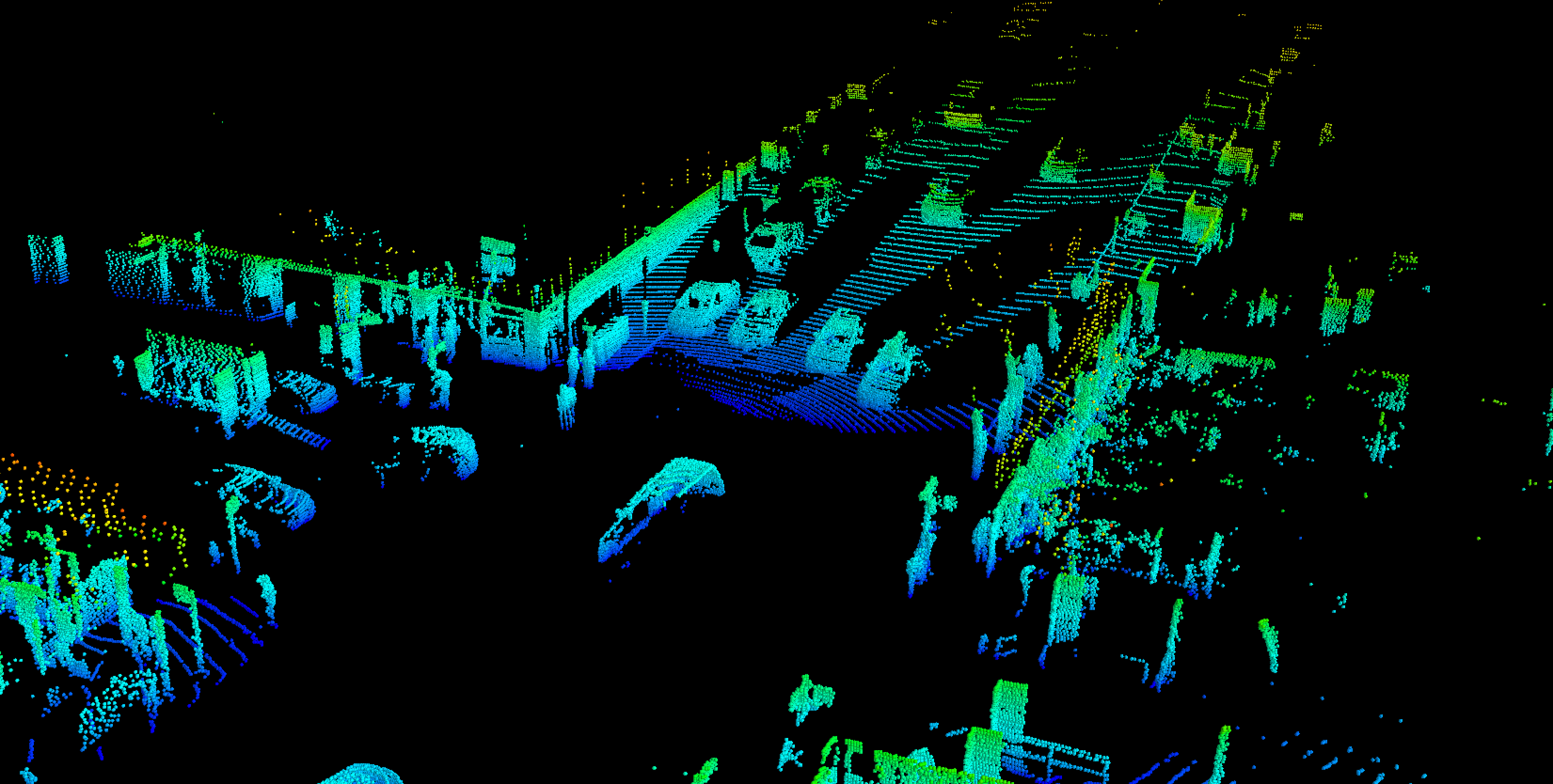}\hspace{0.05cm}
    \includegraphics[width=0.3\textwidth]{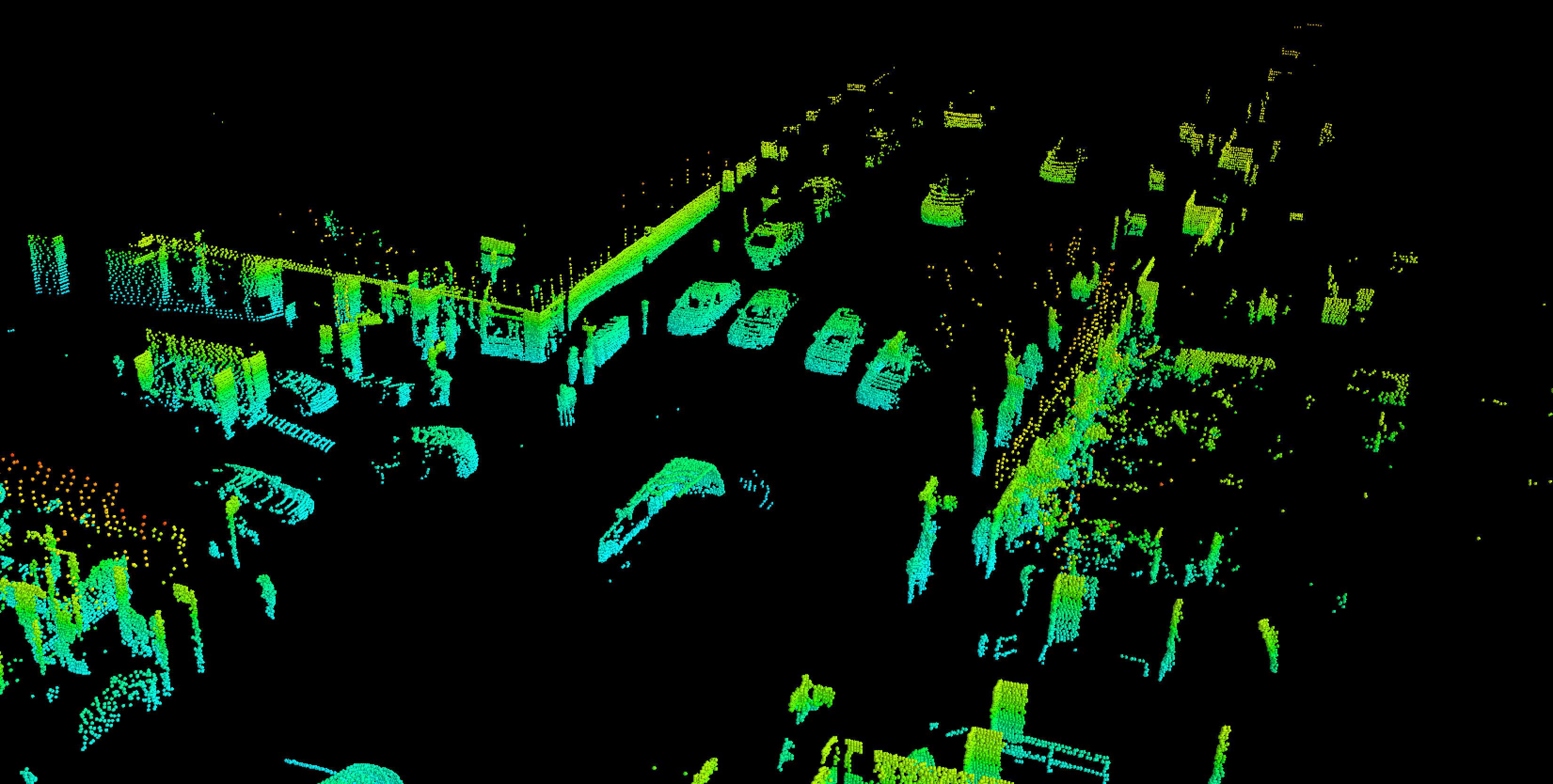}
    \caption{An example of our ground removal technique on a Waymo~\cite{jund2021scalable} scene with a non-planar ground. In all images, color coding indicates the height of each point. From left to right: the input point cloud, the result of thresholding, our result.}
    \label{fig:ground}
\end{figure*}

\textbf{Ground Removal:} The sampling pattern of a LiDAR sensor creates ``swimming'' artifacts as the sensor moves. These artifacts create the appearance of motion and present a large problem for the nearest neighbor loss function used by almost all self-supervised methods. The largest artifacts come from the ground, leading most methods to remove them by height thresholding or by fitting a plane~\cite{baur2021slim,liu2019flownet3d}. This can fail when the ground changes elevation either from hills or even sidewalks, see \cref{fig:ground}. We propose an improvement based on analyzing the assumptions behind the current approaches. These assumptions are:
\begin{enumerate}
    \item The sensor has been calibrated such that the ground can be represented as a height map $h = f(x, y)$.
    \item Except for a small number of noise returns, all measured points $(x, y, z)$ satisfy $z \geq f(x,y)$
    \item The ground function $f(x, y)$ is in some way ``simple''. Existing methods make use of this assumption by requiring $f(x, y) = c$ for thresholding or $f(x, y) = ax + by + c$ for plane fitting.
\end{enumerate}
The issue comes from an overly strict application of assumption (3). Instead, we allow our height map to be \emph{piecewise linear} rather than linear. To represent our piecewise linear height map we use a 3-layer coordinate network with ReLU activations and 64 hidden units per layer. To fit it we use assumption (2) to design a one-sided robust loss:
\begin{equation}
    \mathcal{L}_{height}(h, z) = \begin{cases}
        (h - z)^2 & z < h\\
        \text{Huber}(h, z) & h \leq z
    \end{cases}.
\end{equation}
The Huber function~\cite{huber1992robust} allows the loss to ignore points high above the ground. For each point cloud the parameters of our height-map $\theta_h$ are found by optimizing
\begin{equation}
    \min\limits_{\theta_h} \sum\limits_{i=1}^{N} \mathcal{L}_{height}(f_{\theta_h}(x_i, y_i), z_i).
\end{equation}
Finally, any points which are less than 0.3m above our predicted ground are removed.

\textbf{Scene Flow Estimation:} Once the ego-motion and ground points have been removed, we estimate the scene flow using the method of Li \etal~\cite{li2021neural}. Briefly, this means that we represent the forwards and backward flow using two coordinate networks $f_{\theta_{fw}}, f_{\theta_{bw}} \in \Rl^{N \times 3} \to \Rl^{N \times 3}$ which are optimized with gradient descent on the objective:
\begin{equation}
    \min\limits_{\theta_{fw}, \theta_{bw}} \mathcal{C}(g_{\theta_{fw}}(\P^{t}), \P^{t + \Delta}) + \mathcal{C}(g_{\theta_{bw}}(g_{\theta_{fw}}(\P^t)), \P^{t}).
\end{equation}
For compactness, we have let $g(\X) = f(\X) + \X$ and $\mathcal{C}$ be the truncated symmetric chamfer distance as described in \cite{li2021neural}. We differ from the original method in that we do not use $\ell_2$ regularization on the weights. We find that when combined with motion compensation that regularization leads to zero flow predictions everywhere.

\begin{figure*}
    \centering
    \includegraphics[width=0.9\linewidth, trim=0 0.25 0 0.0cm,clip]{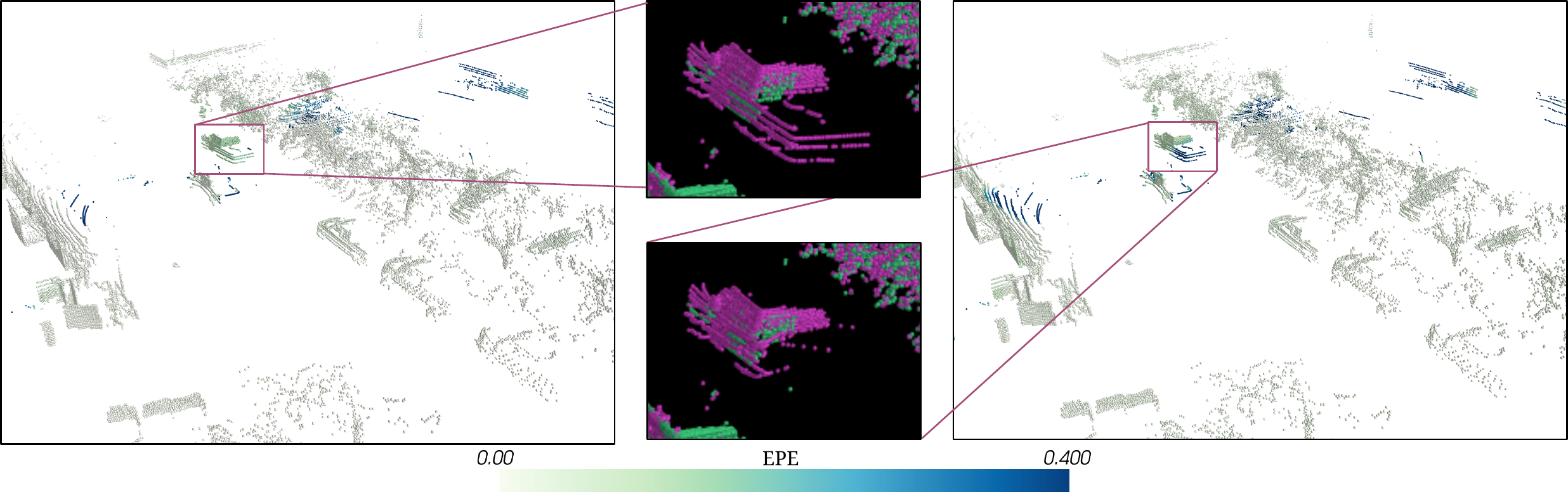}
    \caption{Optimizing for the standard nearest neighbor self-supervised loss can cause mis-predictions in the presence of strong occlusions. Here, the bed of the truck is collapsed into the cab by NSFP (\textbf{left}). Our RANSAC non-rigid refinement step can fix this type of error (\textbf{right}).}
    \label{fig:rigid}
\end{figure*}

\textbf{Rigid Refinement:} Piecewise rigidity has been a common choice of prior for scene flow estimation and has been widely used for LiDAR scene flow~\cite{gojcic2021weakly,dong2022exploiting,baur2021slim}. Many learning-based methods require differentiable rigid refinement for loss functions, but we show that simply fitting rigid motion to the final predictions performs better. This is similar to the method of \cite{gojcic2021weakly} but we use RANSAC to filter outliers. First we use DBSCAN~\cite{ester1996density} to produce a set of clusters $\{\mathcal{V}_j = \{\p_k\}_{k=1}^{K}\}_{j=1}^{J}$. Then for each cluster, we use RANSAC~\cite{fischler1981random} to fit a rigid model to flow predictions for that cluster. That is for each RANSAC iteration we randomly sample 3 points $\p_1, \p_2, \p_3$ (the minimum required to get a unique solution) and their associated flow vectors $\f_1, \f_2, \f_3$ and then fit rigid motion parameters by solving:
\begin{equation}
    \min_{\mathbf{R} \in SO(3), \tb \in \Rl^{3}} \sum_{i=1}^{3} \lVert \mathbf{R}\p_i + \tb - (\p_i + \f_i) \rVert_2^2,
\end{equation}
with the Kabsch algorithm~\cite{kabsch1976solution}. We then compute the norm of the difference between the raw and rigid flows, considering any below a threshold to be inliers. At the end of $T$ iterations the rigid parameters with the highest number of inliers are selected and the parameters are recomputed with respect to the inlier set producing $\mathbf{R}^{*}, \tb^{*}$. Since we know that with motion-compensated inputs the vast majority of flows should be zero, we further refine the flows by setting the rigid motion parameters to the identity transform if $\norm{\tb^*}_2$ is below a threshold. Then all points in the cluster are assigned the flow $\f_i = \mathbf{R}^{*}\p_i + \tb^{*} - \p_i$. Points that were not assigned to any cluster by DBSCAN have their predictions unchanged. The effect of this step can be seen in \cref{fig:rigid}.
\begin{table}
  \setlength{\tabcolsep}{2pt}
  \scriptsize
  \centering
\begin{tabular}{lccccccc}
    \toprule
    & \multirow{3}{*}{Supervision} & \multicolumn{4}{c}{EPE} & AccR & AccS\\
    \cmidrule(lr){3-6}
    & & Avg & Dynamic & \multicolumn{2}{c}{Static} & Dynamic & Dynamic\\
    \cmidrule(lr){5-6}
    & & & FG & FG & BG & FG & FG\\
    \midrule
    Gojcic~\cite{gojcic2021weakly} & Weak & \underline{0.083} & \underline{0.155} & 0.064 & \underline{0.032}  & \underline{0.650} & \underline{0.368}\\
    EgoFlow~\cite{tishchenko2020self} & Weak & 0.205 & 0.447 & 0.079 & 0.090  & 0.111 & 0.018\\
    Sim2Real~\cite{jin2022deformation} & Synth & 0.157 & 0.229 & 0.106 & 0.137  & 0.565 & 0.254\\
    \midrule
    PPWC~\cite{wu2020pointpwc} & Self & 0.130 & 0.168 & 0.092 & 0.129  & 0.556 & 0.229\\
    FlowStep3D~\cite{kittenplon2021flowstep3d} & Self & 0.161 & 0.173 & 0.132 & 0.176  & 0.553 & 0.248\\
    \midrule
    Odometry & None & 0.198 & 0.583 & 0.010 & 0.000  & 0.108 & 0.002\\    
    ICP~\cite{chen1992object} & None & 0.204 & 0.557 & \textbf{0.025} & \textbf{0.028}  & 0.112 & 0.015\\
    NSFP~\cite{li2021neural} & None & 0.088 & 0.193 & \underline{0.033} & 0.039  & 0.542 & 0.327\\
    \textbf{Ours} & None & \textbf{0.055} & \textbf{  
      0.105} & \underline{0.033} & \textbf{0.028} & \textbf{0.777} & \textbf{0.537}\\
    \bottomrule
  \end{tabular}
\caption{Quantitative results on Argoverse 2. Our baseline predicts the motion of dynamic objects by 30\% over \cite{gojcic2021weakly}, despite that method using ground truth foreground masks for training. We also see from the static background EPE that ICP outperforms learning-based methods at predicting ego-motion.}
\label{tab:argoverse}
\end{table}
\begin{table}
  \setlength{\tabcolsep}{2pt}
  \scriptsize
  \centering
\begin{tabular}{lccccccc}
  \toprule
  & \multirow{3}{*}{Supervision} & \multicolumn{4}{c}{EPE} & AccR & AccS\\
  \cmidrule(lr){3-6}
  & & Avg & Dynamic & \multicolumn{2}{c}{Static} & Dynamic & Dynamic\\
  \cmidrule(lr){5-6}
  & & & FG & FG & BG & FG & FG\\
  \midrule
  Gojcic~\cite{gojcic2021weakly} & Weak & 0.084  & 0.145 & 0.060 & 0.047 & 0.714 & 0.436\\
  EgoFlow~\cite{tishchenko2020self} & Weak & 0.236 & 0.520 & 0.074 & 0.114 & 0.105 & 0.022\\
  Sim2Real~\cite{jin2022deformation} & Synth & 0.214 & 0.305 & 0.143 & 0.194 & 0.426 & 0.155\\
  \midrule
  PPWC~\cite{wu2020pointpwc} & Self & 0.156 & 0.197 & 0.097 & 0.173 & 0.468 & 0.183\\
  FlowStep3D~\cite{kittenplon2021flowstep3d} & Self & 0.156 & 0.155 & 0.090 & 0.224 & 0.660 & 0.370\\
  \midrule
  ICP &  None & 0.176 & 0.450 & 0.033 & \textbf{0.046} & 0.151 & 0.047\\
  NSFP & None & 0.088 & 0.130 & 0.034 & 0.101 & 0.711 & 0.447\\
  \textbf{Ours} & None & \textbf{0.055} & \textbf{0.061} & \textbf{0.020} & 0.083 & \textbf{0.891} & \textbf{0.681}\\
  \bottomrule
\end{tabular}

\caption{Quantitative results on NuScenes. Our baseline halves the EPE on dynamic objects when compared to the next best method (NSFP). Since our baseline also uses NSFP as its backbone, this indicates that pre- and post-processing steps can have an enormous impact. Again we also see that ICP has the best ego-motion prediction.}
\label{tab:nuscenes}
\end{table}
\begin{table}
  \setlength{\tabcolsep}{2pt}
  \scriptsize
  \centering
\begin{tabular}{lccccccc}
  \toprule
  & \multirow{3}{*}{Supervision} & \multicolumn{4}{c}{EPE} & AccR & AccS\\
  \cmidrule(lr){3-6}
  & & Avg & Dynamic & \multicolumn{2}{c}{Static} & Dynamic & Dynamic\\
  \cmidrule(lr){5-6}
  & & & FG & FG & BG & FG & FG\\
  \midrule
  Gojcic~\cite{gojcic2021weakly} & Weak & 0.059 & 0.108 & 0.045 & \textbf{0.025} & 0.844 & 0.584\\
  EgoFlow~\cite{tishchenko2020self} & Weak & 0.183 & 0.390 & 0.069 & 0.089 & 0.178 & 0.046\\
  Sim2Real~\cite{jin2022deformation} & Synth & 0.132 & 0.180 & 0.075 & 0.142 & 0.497 & 0.217\\
  \midrule
  PPWC~\cite{wu2020pointpwc} & Self & 0.132 & 0.180 & 0.075 & 0.142 & 0.497 & 0.217\\
  FlowStep3D~\cite{kittenplon2021flowstep3d} & Self & 0.169 & 0.152 & 0.123 & 0.232 & 0.708 & 0.405\\
  \midrule
  ICP & None & 0.192 & 0.498 & \underline{0.022} & 0.055 & 0.172 & 0.047\\
  NSFP & None & 0.100 & 0.171 & 0.021 & 0.108 & 0.539 & 0.331\\
  \textbf{Ours} & None & \textbf{0.041} & \textbf{0.073} & \textbf{0.013} & 0.039 & \textbf{0.877} & \textbf{0.726}\\
  \bottomrule
\end{tabular}
\caption{Quantitative results on Waymo. Our baseline outperforms all tested methods but notably in this instance \cite{gojcic2021weakly} performs better than ICP at predicting the background ego-motion.}
\label{tab:waymo}
\end{table}
\begin{table}
\scriptsize
    \centering
    \begin{tabular}{lcccc}
    \toprule
         & \multicolumn{2}{c}{Moving} & Static & 50-50\\
         \cmidrule(lr){2-3}
         & EPE & Accuracy Relax & EPE & EPE \\\midrule
         Zero & 0.6381 & 0.1632 & 0.5248 & 0.5814\\
         ICP & 0.2101 & 0.6151 & \textbf{0.0290} & 0.1196\\
         PPWC & 0.3539 & 0.2543 & 0.1974 & 0.2756\\
         EgoFlow & 0.7399 & 0.0000 & 0.0570 & 0.3985\\
         SLIM (U) & 0.1050 & 0.7365 & 0.0925 & 0.0987\\
         Ours & \textbf{0.0625} & \textbf{0.894} & 0.0660 & \underline{0.064}\\
         \midrule
         SLIM (S) & \underline{0.0702} &  \underline{0.8921} & \underline{0.0499} & \textbf{0.0600}\\
         \bottomrule
    \end{tabular}
    \caption{The results of our baseline on NuScenes as evaluated by Baur \etal~\cite{baur2021slim}. U and S refer to the unsupervised and fully-supervised versions of their method. Our dataless baseline outperforms all self-supervised methods and even achieves comparable and superior results to the fully supervised network.}
    \label{tab:slim_comp}
\end{table}
\begin{table}
\scriptsize
    \centering
    \begin{tabular}{lcccc}
    \toprule
    & Supervision & EPE & AccS & AccR\\\midrule
    PPWC~\cite{wu2020pointpwc} & Full & 0.710 & 0.114 & 0.219\\\
    FLOT~\cite{puy2020flot} & Full & 0.773 & 0.084 & 0.177\\
    MeteorNet~\cite{liu2019meteornet} & Full & 0.277 & / & /\\\midrule
    Gojcic~\cite{gojcic2021weakly} & Weak & 0.133 & 0.460 & 0.746\\
    Gojcic++ & Weak & 0.102 & 0.686 & 0.819\\
    Dong (Waymo Open)~\cite{dong2022exploiting} & Weak & 0.077 & 0.812 & 0.906\\
    Dong (Semantic KITTI) & Weak & \underline{0.065} & \underline{0.857} & \underline{0.940}\\\midrule
    Ours & None & \textbf{0.061} & \textbf{0.917} & \textbf{0.962}\\
    \bottomrule
    \end{tabular}
    \caption{Results of our method on lidarKITTI w/ ground \cite{gojcic2021weakly}. We outperform all existing methods without using any training data.}
    \label{tab:lidar-kitti}
\end{table}
\section{Evaluation}

Our central claim is that the use of popular benchmarks causes leading methods to degrade in quality when evaluated on real-world data. The main result in support of this is shown in \cref{fig:perf-over-time}; here we detail the procedure used to select, train, and evaluate the tested methods (\cref{sec:benchmark}). We also discuss the performance of our baseline, which we find outperforms all the tested methods, as well as validate these results through comparison to other authors's self-reported metrics on NuScenes and lidarKITTI (\cref{sec:existing}). Finally, we evaluate our ground segmentation method (\cref{sec:ground}). 
\subsection{Training and Evaluation Protocol}
\label{sec:benchmark}
\textbf{Evaluation Metrics:} We use a set of standard metrics:
\begin{itemize}
    \item \textbf{EPE:} Average end-point-error \ie the $\ell_2$ norm of the difference between predicted and ground truth flow.
    \item \textbf{Accuracy Relax:} Ratio of predictions with absolute EPE less than 0.1m or relative error below 0.1.
    \item \textbf{Accuracy Strict:} Same as Accuracy Relax but with a threshold of 0.05.
\end{itemize}

Rather than averaging these metrics over the entire dataset, we break them into three classes of points: dynamic foreground, static foreground, and static background. Points are considered belonging to the foreground if they are contained in the bounding box of some tracked object and they are considered dynamic if they have a flow magnitude of at least $\SI{0.5}{\metre\per\second}$. The choice of threshold is discussed in the supplement. Separating the metrics in this way is vital due to the low ratio of dynamic to static points, without it the metrics are dominated by performance on the static background. In order to produce a single number for ranking purposes we combine the EPE results on the classes into a three-way average similar to \cite{baur2021slim}. In order to save computation we evaluate on a subset on of each dataset's validation split. We use the entire NuScenes validation set, and every 5th frame of the Argoverse and Waymo validation sets.

\textbf{Method Selection:} We chose to evaluate 6 methods to serve as a thorough exploration of recent research. Methods were chosen if they presented a self or weakly supervised method, appeared in a recent conference, and made available a working implementation of their method. This led us to choose: PointPWC Net~\cite{wu2020pointpwc}, EgoFlow~\cite{tishchenko2020self}, FlowStep3D~\cite{kittenplon2021flowstep3d}, NSFP~\cite{li2021neural}, Sim2Real~\cite{jin2022deformation}, and Gojcic \etal~\cite{gojcic2021weakly}. Additionally, we include as baselines an off-the-shelf ICP~\cite{beaton1974fitting} implementation~\cite{vizzo2023ral}. Most of the baseline methods clip points to a depth of 35m~\cite{wu2020pointpwc,jin2022deformation,tishchenko2020self,kittenplon2021flowstep3d}. To make comparisons fair we only include points contained in a 70m square around the sensor as was done in \cite{baur2021slim}.

\textbf{Training Procedure:} As much as possible we attempted to use the same training strategy as the original authors, but we also wished to enforce standardization on the amount of computing resources allocated to each method. As a result, we chose to adopt the two-stage training regime from \cite{wu2020pointpwc}. First, we train for seven days on a quarter of the whole training set followed by five days of fine-tuning on the entire dataset. Each method was trained using the largest batch size possible on a single NVIDIA T4 GPU (some methods were not set up for multi-GPU training), and using the authors' optimizer and learning rate schedule. Methods that were able to include ground truth ego-motion in their loss formulation were also given that information. We also used ground truth foreground/background masks to train the weakly supervised method \cite{gojcic2021weakly}. The only method that required substantial changes was \cite{jin2022deformation} as its training is based on transferring information from a synthetic labeled dataset. The synthetic data comes from a virtual depth camera, so we clip the LiDAR points to match the field of view.

\textbf{Results:} Our results can be found in tables \ref{tab:argoverse}, \ref{tab:nuscenes}, and \ref{tab:waymo}. The main result in \cref{fig:perf-over-time} was created by taking the dynamic foreground error for each learning method/dataset combination and plotting it against their self-reported stereoKITTI error. We use the dynamic foreground EPE since this is the hardest and most important component of the real-world scene flow problem. We remove EgoFlow from these plots since it is a large outlier in terms of both real-world dynamic EPE and stereoKITTI EPE.

We also claim that synthetic benchmarks are causing researchers to focus on the wrong problems, which we test by comparing the tested methods to our simple test-time optimization baseline. Validating this claim, we can observe that our baseline outperforms all tested methods despite not using any training data; even Gojcic \etal who use ground truth foreground masks. Each tested method claims some learning-based novelty as its main contribution, which is then shown to be effective through experiments on KITTI-SF and FlyingThings3D. However, none can outperform a carefully designed baseline when evaluated on real data. Further validating this claim is the fact that no method was able to match the performance of ICP at predicting the ego-motion. This is in spite of the fact that several of the methods~\cite{gojcic2021weakly,baur2021slim,tishchenko2020self} explicitly claim ego-motion predictions as a benefit of their architectures. As discussed in \cref{sec:dataset-problems}, this results from the unrealistic dynamic ratio found in the standard benchmarks. Gojcic \etal comes the closest to outperforming ICP and does so on Waymo. However, this is because it essentially incorporates ICP as a part of its final test-time refinement of the ego-motion. 


\subsection{Existing Benchmark Comparison}
\label{sec:existing}
There may be some concern that the tested methods performed poorly compared to the baseline due to a lack of tuning of hyperparameters. To address this concern, we also compare our baseline to self-reported metrics on two real-world datsets: NuScenes as evaluted in \cite{baur2021slim} and lidarKITTI which was  introduced by the Gojcic \etal ~\cite{gojcic2021weakly}.

\textbf{NuScenes:} The authors of SLIM~\cite{baur2021slim} also reported numbers on real-world data. As can be seen in \cref{tab:slim_comp} our method vastly outperforms SLIM and the other methods they tested on the main 50-50 EPE metric, dynamic EPE, and dynamic accuracy. Again, ICP performs far better than every other method on static points. We also include the SLIM results when trained in a supervised manner and show that we achieve comparable and some superior results, despite not using any training data.

\textbf{lidarKITTI:}
lidarKITTI~\cite{gojcic2021weakly} was generated by associating KITTI-SF labels with the raw LiDAR scans. We use the version with ground points as ground removal is a component of our method. Ground point flow is set to the ego-motion. The results are shown in \cref{tab:lidar-kitti} and further confirm that our baseline performs well even compared to self-reported numbers. It substantially outperforms all but one of the existing self- and weakly-supervised methods.

\subsection{Ground Segmentation}
\label{sec:ground}
We close with an evaluation of our ground segmentation method. First, we present a qualitative example (\cref{fig:ground}) showing our method effectively handling a scene with a non-planar ground. To quantitatively analyze our method we need to consider what constitutes failure. The subtlety lies in the ill-defined nature of the ground, making it impossible to define normal precision and recall metrics. Given that our goal is motion estimation, the failure we are most concerned with is classifying a dynamic point as belonging to the ground. Therefore we look at the rate at which we make this error. On NuScenes we find that 99.3\% of the time points which are classified as ground are in fact static and achieve a rate of 99.4\% on Waymo.
\section{Conclusion}
We re-examined the evaluations of self- and weakly- supervised scene flow methods in the context of autonomous driving and found several deficiencies. We claimed that these deficiencies are impacting the quality and types of recently proposed methods. To provide evidence for this claim we evaluated a large number of top methods on several real-world datasets. We found a negative correlation between performance on the standard stereoKITTI benchmark and performance on all of the real-world datasets. Additionally, we proposed a dataless estimation technique that far outperformed the existing approaches, demonstrating that focus on the current benchmarks is causing researchers to ignore effective methods. Given that our baseline is based on pre- and post-processing techniques, we believe that other methods not based on test-time optimization will also benefit from them.
{\small
\bibliographystyle{ieee_fullname}
\bibliography{egbib}

\begin{thebibliography}{10}\itemsep=-1pt

\bibitem{amberg2007optimal}
Brian Amberg, Sami Romdhani, and Thomas Vetter.
\newblock Optimal step nonrigid icp algorithms for surface registration.
\newblock In {\em IEEE Conf. Comput. Vis. Pattern Recog.}, pages 1--8. IEEE,
  2007.

\bibitem{atzmon2020sal}
Matan Atzmon and Yaron Lipman.
\newblock Sal: Sign agnostic learning of shapes from raw data.
\newblock In {\em IEEE Conf. Comput. Vis. Pattern Recog.}, pages 2565--2574,
  2020.

\bibitem{basha2013multi}
Tali Basha, Yael Moses, and Nahum Kiryati.
\newblock Multi-view scene flow estimation: A view centered variational
  approach.
\newblock {\em Int. J. Comput. Vis.}, 101(1):6--21, 2013.

\bibitem{baur2021slim}
Stefan~Andreas Baur, David~Josef Emmerichs, Frank Moosmann, Peter Pinggera,
  Bj{\"o}rn Ommer, and Andreas Geiger.
\newblock Slim: Self-supervised lidar scene flow and motion segmentation.
\newblock In {\em Int. Conf. Comput. Vis.}, pages 13126--13136, 2021.

\bibitem{beaton1974fitting}
Albert~E Beaton and John~W Tukey.
\newblock The fitting of power series, meaning polynomials, illustrated on
  band-spectroscopic data.
\newblock {\em Technometrics}, 16(2):147--185, 1974.

\bibitem{behl2019pointflownet}
Aseem Behl, Despoina Paschalidou, Simon Donn{\'e}, and Andreas Geiger.
\newblock Pointflownet: Learning representations for rigid motion estimation
  from point clouds.
\newblock In {\em Int. Conf. Comput. Vis.}, pages 7962--7971, 2019.

\bibitem{caesar2020nuscenes}
Holger Caesar, Varun Bankiti, Alex~H Lang, Sourabh Vora, Venice~Erin Liong,
  Qiang Xu, Anush Krishnan, Yu Pan, Giancarlo Baldan, and Oscar Beijbom.
\newblock nuscenes: A multimodal dataset for autonomous driving.
\newblock In {\em IEEE Conf. Comput. Vis. Pattern Recog.}, pages 11621--11631,
  2020.

\bibitem{chen1992object}
Yang Chen and G{\'e}rard Medioni.
\newblock Object modelling by registration of multiple range images.
\newblock {\em Img. Vis. Comput.}, 10(3):145--155, 1992.

\bibitem{chen2019learning}
Zhiqin Chen and Hao Zhang.
\newblock Learning implicit fields for generative shape modeling.
\newblock In {\em IEEE Conf. Comput. Vis. Pattern Recog.}, pages 5939--5948,
  2019.

\bibitem{chui2003new}
Haili Chui and Anand Rangarajan.
\newblock A new point matching algorithm for non-rigid registration.
\newblock {\em Comput. Vis. Img. Und.}, 89(2-3):114--141, 2003.

\bibitem{dewan2016rigid}
Ayush Dewan, Tim Caselitz, Gian~Diego Tipaldi, and Wolfram Burgard.
\newblock Rigid scene flow for 3d lidar scans.
\newblock In {\em Int. Conf. Intel. Rob. Sys.}, pages 1765--1770. IEEE, 2016.

\bibitem{dong2022exploiting}
Guanting Dong, Yueyi Zhang, Hanlin Li, Xiaoyan Sun, and Zhiwei Xiong.
\newblock Exploiting rigidity constraints for lidar scene flow estimation.
\newblock In {\em IEEE Conf. Comput. Vis. Pattern Recog.}, pages 12776--12785,
  2022.

\bibitem{eisenberger2020smooth}
Marvin Eisenberger, Zorah Lahner, and Daniel Cremers.
\newblock Smooth shells: Multi-scale shape registration with functional maps.
\newblock In {\em IEEE Conf. Comput. Vis. Pattern Recog.}, pages 12265--12274,
  2020.

\bibitem{ester1996density}
Martin Ester, Hans-Peter Kriegel, J\"{o}rg Sander, and Xiaowei Xu.
\newblock A density-based algorithm for discovering clusters in large spatial
  databases with noise.
\newblock In {\em Int. Conf. Know. Disc. Data Min.}, KDD'96, page 226–231.
  AAAI Press, 1996.

\bibitem{fischler1981random}
Martin~A Fischler and Robert~C Bolles.
\newblock Random sample consensus: a paradigm for model fitting with
  applications to image analysis and automated cartography.
\newblock {\em Comm. ACM}, 24(6):381--395, 1981.

\bibitem{geiger2012we}
Andreas Geiger, Philip Lenz, and Raquel Urtasun.
\newblock Are we ready for autonomous driving? the kitti vision benchmark
  suite.
\newblock In {\em IEEE Conf. Comput. Vis. Pattern Recog.}, pages 3354--3361.
  IEEE, 2012.

\bibitem{gojcic2021weakly}
Zan Gojcic, Or Litany, Andreas Wieser, Leonidas~J Guibas, and Tolga Birdal.
\newblock Weakly supervised learning of rigid 3d scene flow.
\newblock In {\em IEEE Conf. Comput. Vis. Pattern Recog.}, pages 5692--5703,
  2021.

\bibitem{gu2019hplflownet}
Xiuye Gu, Yijie Wang, Chongruo Wu, Yong~Jae Lee, and Panqu Wang.
\newblock Hplflownet: Hierarchical permutohedral lattice flownet for scene flow
  estimation on large-scale point clouds.
\newblock In {\em IEEE Conf. Comput. Vis. Pattern Recog.}, pages 3254--3263,
  2019.

\bibitem{hadfield2011kinecting}
Simon Hadfield and Richard Bowden.
\newblock Kinecting the dots: Particle based scene flow from depth sensors.
\newblock In {\em Int. Conf. Comput. Vis.}, pages 2290--2295. IEEE, 2011.

\bibitem{hadfield2013scene}
Simon Hadfield and Richard Bowden.
\newblock Scene particles: Unregularized particle-based scene flow estimation.
\newblock {\em IEEE Trans. Pattern Anal. Mach. Intell.}, 36(3):564--576, 2013.

\bibitem{himmelsbach2010fast}
Michael Himmelsbach, Felix~V Hundelshausen, and H-J Wuensche.
\newblock Fast segmentation of 3d point clouds for ground vehicles.
\newblock In {\em Intel. Veh. Symp. IV}, pages 560--565. IEEE, 2010.

\bibitem{hornacek2014sphereflow}
Michael Hornacek, Andrew Fitzgibbon, and Carsten Rother.
\newblock Sphereflow: 6 dof scene flow from rgb-d pairs.
\newblock In {\em IEEE Conf. Comput. Vis. Pattern Recog.}, pages 3526--3533,
  2014.

\bibitem{huber1992robust}
Peter~J Huber.
\newblock Robust estimation of a location parameter.
\newblock In {\em Breakthroughs in statistics}, pages 492--518. Springer, 1992.

\bibitem{izadi2011kinectfusion}
Shahram Izadi, David Kim, Otmar Hilliges, David Molyneaux, Richard Newcombe,
  Pushmeet Kohli, Jamie Shotton, Steve Hodges, Dustin Freeman, Andrew Davison,
  and Andrew Fitzgibbon.
\newblock Kinectfusion: Real-time 3d reconstruction and interaction using a
  moving depth camera.
\newblock In {\em User Int. Soft. Tech.}, '11, page 559–568, New York, NY,
  USA, 2011. Association for Computing Machinery.

\bibitem{jimenez2021ground}
V{\'\i}ctor Jim{\'e}nez, Jorge Godoy, Antonio Artu{\~n}edo, and Jorge Villagra.
\newblock Ground segmentation algorithm for sloped terrain and sparse lidar
  point cloud.
\newblock {\em IEEE Access}, 9:132914--132927, 2021.

\bibitem{jin2022deformation}
Zhao Jin, Yinjie Lei, Naveed Akhtar, Haifeng Li, and Munawar Hayat.
\newblock Deformation and correspondence aware unsupervised synthetic-to-real
  scene flow estimation for point clouds.
\newblock In {\em IEEE Conf. Comput. Vis. Pattern Recog.}, pages 7233--7243,
  2022.

\bibitem{jund2021scalable}
Philipp Jund, Chris Sweeney, Nichola Abdo, Zhifeng Chen, and Jonathon Shlens.
\newblock Scalable scene flow from point clouds in the real world.
\newblock {\em IEEE Rob. Aut. Letters}, 7(2):1589--1596, 2021.

\bibitem{kabsch1976solution}
Wolfgang Kabsch.
\newblock A solution for the best rotation to relate two sets of vectors.
\newblock {\em Acta Crystallographica Section A: Crystal Physics, Diffraction,
  Theoretical and General Crystallography}, 32(5):922--923, 1976.

\bibitem{kingma2014adam}
Diederik~P Kingma and Jimmy Ba.
\newblock Adam: A method for stochastic optimization.
\newblock {\em arXiv preprint arXiv:1412.6980}, 2014.

\bibitem{kittenplon2021flowstep3d}
Yair Kittenplon, Yonina~C Eldar, and Dan Raviv.
\newblock Flowstep3d: Model unrolling for self-supervised scene flow
  estimation.
\newblock In {\em IEEE Conf. Comput. Vis. Pattern Recog.}, pages 4114--4123,
  2021.

\bibitem{lee2022patchwork++}
Seungjae Lee, Hyungtae Lim, and Hyun Myung.
\newblock Patchwork++: Fast and robust ground segmentation solving partial
  under-segmentation using 3d point cloud.
\newblock In {\em Int. Conf. Intel. Rob. Sys.}, pages 13276--13283. IEEE, 2022.

\bibitem{li2008global}
Hao Li, Robert~W Sumner, and Mark Pauly.
\newblock Global correspondence optimization for non-rigid registration of
  depth scans.
\newblock In {\em Comput. Graph. For.}, volume~27, pages 1421--1430. Wiley
  Online Library, 2008.

\bibitem{li2022rigidflow}
Ruibo Li, Chi Zhang, Guosheng Lin, Zhe Wang, and Chunhua Shen.
\newblock Rigidflow: Self-supervised scene flow learning on point clouds by
  local rigidity prior.
\newblock In {\em IEEE Conf. Comput. Vis. Pattern Recog.}, pages 16959--16968,
  2022.

\bibitem{li2021neural}
Xueqian Li, Jhony Kaesemodel~Pontes, and Simon Lucey.
\newblock Neural scene flow prior.
\newblock {\em Adv. Neural Inform. Process. Syst.}, 34:7838--7851, 2021.

\bibitem{liu2019flownet3d}
Xingyu Liu, Charles~R Qi, and Leonidas~J Guibas.
\newblock Flownet3d: Learning scene flow in 3d point clouds.
\newblock In {\em IEEE Conf. Comput. Vis. Pattern Recog.}, pages 529--537,
  2019.

\bibitem{liu2019meteornet}
Xingyu Liu, Mengyuan Yan, and Jeannette Bohg.
\newblock Meteornet: Deep learning on dynamic 3d point cloud sequences.
\newblock In {\em IEEE Conf. Comput. Vis. Pattern Recog.}, pages 9246--9255,
  2019.

\bibitem{mayer2016large}
Nikolaus Mayer, Eddy Ilg, Philip Hausser, Philipp Fischer, Daniel Cremers,
  Alexey Dosovitskiy, and Thomas Brox.
\newblock A large dataset to train convolutional networks for disparity,
  optical flow, and scene flow estimation.
\newblock In {\em IEEE Conf. Comput. Vis. Pattern Recog.}, pages 4040--4048,
  2016.

\bibitem{menze2015object}
Moritz Menze and Andreas Geiger.
\newblock Object scene flow for autonomous vehicles.
\newblock In {\em Int. Conf. Comput. Vis.}, pages 3061--3070, 2015.

\bibitem{menze2015joint}
Moritz Menze, Christian Heipke, and Andreas Geiger.
\newblock Joint 3d estimation of vehicles and scene flow.
\newblock {\em ISPRS Annals Photo. Rem. Sens. Spat. Inf. Sci.}, 2:427, 2015.

\bibitem{mescheder2019occupancy}
Lars Mescheder, Michael Oechsle, Michael Niemeyer, Sebastian Nowozin, and
  Andreas Geiger.
\newblock Occupancy networks: Learning 3d reconstruction in function space.
\newblock In {\em IEEE Conf. Comput. Vis. Pattern Recog.}, pages 4460--4470,
  2019.

\bibitem{Mittal_2020_CVPR}
Himangi Mittal, Brian Okorn, and David Held.
\newblock Just go with the flow: Self-supervised scene flow estimation.
\newblock In {\em IEEE Conf. Comput. Vis. Pattern Recog.}, June 2020.

\bibitem{moosmann2009segmentation}
Frank Moosmann, Oliver Pink, and Christoph Stiller.
\newblock Segmentation of 3d lidar data in non-flat urban environments using a
  local convexity criterion.
\newblock In {\em Intel. Veh. Symp. IV}, pages 215--220. IEEE, 2009.

\bibitem{najibi2022motion}
Mahyar Najibi, Jingwei Ji, Yin Zhou, Charles~R Qi, Xinchen Yan, Scott Ettinger,
  and Dragomir Anguelov.
\newblock Motion inspired unsupervised perception and prediction in autonomous
  driving.
\newblock In {\em Eur. Conf. Comput. Vis.}, pages 424--443. Springer, 2022.

\bibitem{narksri2018slope}
Patiphon Narksri, Eijiro Takeuchi, Yoshiki Ninomiya, Yoichi Morales, Naoki
  Akai, and Nobuo Kawaguchi.
\newblock A slope-robust cascaded ground segmentation in 3d point cloud for
  autonomous vehicles.
\newblock In {\em Int. Conf. Intel. Trans. Sys.}, pages 497--504. IEEE, 2018.

\bibitem{park2019deepsdf}
Jeong~Joon Park, Peter Florence, Julian Straub, Richard Newcombe, and Steven
  Lovegrove.
\newblock Deepsdf: Learning continuous signed distance functions for shape
  representation.
\newblock In {\em IEEE Conf. Comput. Vis. Pattern Recog.}, pages 165--174,
  2019.

\bibitem{pauly2005example}
Mark Pauly, Niloy~J Mitra, Joachim Giesen, Markus~H Gross, and Leonidas~J
  Guibas.
\newblock Example-based 3d scan completion.
\newblock In {\em Symp. Geom. Proc.}, pages 23--32, 2005.

\bibitem{pons2007multi}
Jean-Philippe Pons, Renaud Keriven, and Olivier Faugeras.
\newblock Multi-view stereo reconstruction and scene flow estimation with a
  global image-based matching score.
\newblock {\em Int. J. Comput. Vis.}, 72(2):179--193, 2007.

\bibitem{pons2003variational}
J-P Pons, Renaud Keriven, O Faugeras, and Gerardo Hermosillo.
\newblock Variational stereovision and 3d scene flow estimation with
  statistical similarity measures.
\newblock In {\em Int. Conf. Comput. Vis.}, volume~2, pages 597--597. IEEE
  Computer Society, 2003.

\bibitem{pontes2020scene}
Jhony~Kaesemodel Pontes, James Hays, and Simon Lucey.
\newblock Scene flow from point clouds with or without learning.
\newblock In {\em Int. Conf. 3D Vis.}, pages 261--270. IEEE, 2020.

\bibitem{puy2020flot}
Gilles Puy, Alexandre Boulch, and Renaud Marlet.
\newblock Flot: Scene flow on point clouds guided by optimal transport.
\newblock In {\em Eur. Conf. Comput. Vis.}, pages 527--544. Springer, 2020.

\bibitem{qi2017pointnet}
Charles~R Qi, Hao Su, Kaichun Mo, and Leonidas~J Guibas.
\newblock Pointnet: Deep learning on point sets for 3d classification and
  segmentation.
\newblock In {\em IEEE Conf. Comput. Vis. Pattern Recog.}, pages 652--660,
  2017.

\bibitem{sitzmann2020implicit}
Vincent Sitzmann, Julien Martel, Alexander Bergman, David Lindell, and Gordon
  Wetzstein.
\newblock Implicit neural representations with periodic activation functions.
\newblock {\em Adv. Neural Inform. Process. Syst.}, 33:7462--7473, 2020.

\bibitem{tishchenko2020self}
Ivan Tishchenko, Sandro Lombardi, Martin~R Oswald, and Marc Pollefeys.
\newblock Self-supervised learning of non-rigid residual flow and ego-motion.
\newblock In {\em Int. Conf. 3D Vis.}, pages 150--159. IEEE, 2020.

\bibitem{vedula1999three}
Sundar Vedula, Simon Baker, Peter Rander, Robert Collins, and Takeo Kanade.
\newblock Three-dimensional scene flow.
\newblock In {\em IEEE Conf. Comput. Vis. Pattern Recog.}, volume~2, pages
  722--729. IEEE, 1999.

\bibitem{vedula1999}
Sundar Vedula, Simon Baker, Peter Rander, Robert Collins, and Takeo Kanade.
\newblock Three-dimensional scene flow.
\newblock In {\em Int. Conf. Comput. Vis.}, volume~2, pages 722--729. IEEE,
  1999.

\bibitem{vizzo2023ral}
Ignacio Vizzo, Tiziano Guadagnino, Benedikt Mersch, Louis Wiesmann, Jens
  Behley, and Cyrill Stachniss.
\newblock {KISS-ICP: In Defense of Point-to-Point ICP -- Simple, Accurate, and
  Robust Registration If Done the Right Way}.
\newblock {\em IEEE Robotics and Automation Letters (RA-L)}, 8(2):1029--1036,
  2023.

\bibitem{vogel20113d}
Christoph Vogel, Konrad Schindler, and Stefan Roth.
\newblock 3d scene flow estimation with a rigid motion prior.
\newblock In {\em Int. Conf. Comput. Vis.}, pages 1291--1298. IEEE, 2011.

\bibitem{vogel2013piecewise}
Christoph Vogel, Konrad Schindler, and Stefan Roth.
\newblock Piecewise rigid scene flow.
\newblock In {\em Int. Conf. Comput. Vis.}, pages 1377--1384, 2013.

\bibitem{wilson2021argoverse}
Benjamin Wilson, William Qi, Tanmay Agarwal, John Lambert, Jagjeet Singh,
  Siddhesh Khandelwal, Bowen Pan, Ratnesh Kumar, Andrew Hartnett,
  Jhony~Kaesemodel Pontes, et~al.
\newblock Argoverse 2: Next generation datasets for self-driving perception and
  forecasting.
\newblock In {\em Neur. Inform. Process. Syst. Data. Bench. Track}, 2021.

\bibitem{wu2020pointpwc}
Wenxuan Wu, Zhi~Yuan Wang, Zhuwen Li, Wei Liu, and Li Fuxin.
\newblock Pointpwc-net: Cost volume on point clouds for (self-) supervised
  scene flow estimation.
\newblock In {\em Eur. Conf. Comput. Vis.}, pages 88--107. Springer, 2020.

\bibitem{zermas2017fast}
Dimitris Zermas, Izzat Izzat, and Nikolaos Papanikolopoulos.
\newblock Fast segmentation of 3d point clouds: A paradigm on lidar data for
  autonomous vehicle applications.
\newblock In {\em Int. Conf. Rob. Aut.}, pages 5067--5073. IEEE, 2017.

\end{thebibliography}
}
\clearpage
\appendix
\section{Expected Number of Correspondences}
In Sec. 3 of the main paper, we point out that randomly subsampling stereoKITTI's input point clouds does not fix the one-to-one correspondence issue because there are still an expected 745 points with correspondences. Here, we explain how we arrived at that number. Each example in stereoKITTI comprises approximately 90,000 pairs of points $(p_i, q_i)$. Many methods uniformly sample 8192 points from the $p_i$'s, and the $q_i$'s separately. For each chosen $p_i$ there is some probability that its corresponding $q_i$ is also chosen as input, and there is a random variable $X$, which is the number of times this happens. We are interested in computing the expectation of this random variable. Let $\Phi$ be the indices of the 8192 points chosen for the first frame. Now let $x_i$ be an indicator random variable that is 1 if $q_i$ is chosen as input and 0 otherwise. Observe that $X = \sum_{i \in \Phi} x_i$ and $\mathbb{E}[x_i] = p(q_i \text{ is chosen}) = \frac{8192}{90000}$. By the linearity of expectation: $\mathbb{E}[X] = \sum_{i \in \Phi} \mathbb{E}[x_i] = 8192 \cdot \frac{8192}{90000} \approx 745$.
\section{Label Creation}
Multiple authors have used ground truth multi-object tracks to scene-flow labels\cite{baur2021slim,jund2021scalable}. We follow the same procedure to create labels for Argoverse 2.0.

\textbf{Problem Statement:} We assume as input two point clouds $\P^{t} \in \Rl^{N \times 3}, \P^{t+\Delta} \in \Rl^{M \times 3}$. The superscript indicates that these point clouds are separated by a small time delta (usually $\Delta$ = 0.1s), and the variables $N, M$ indicate that the two point clouds have different numbers of points. The goal is to predict a set of flow vectors $\{\f_i \in \Rl^{3}\}_{i=1}^{N}$ which describe the motion of each point from time $t$ to time $t + \Delta$. Some datasets also give access to the ego-motion of the sensor since in the autonomous vehicle setting this information is available from odometry or GPS.

\textbf{Flow Label Creation:} For each point cloud in a MOT dataset we have a set of oriented bounding boxes $\{B_i^{t}\}_{i=1}^{K}$. For each $B_i^{t}$, if the second frame at time $t+\Delta$ contains a corresponding bounding box $B^{t+\Delta}_j$, we can extract the rigid transformation $\R_i^{t}, \tb_i^{t}$ that transforms points in the first box to the second. For each point $\p_j$ inside the bounding box we assign it the flow $\f_j = \R_i\p_j + \tb_i$. Points not belonging to any bounding box are assigned the ego-motion as flow. For objects which only appear in one frame, we cannot compute the ground truth flow and so they are ignored for evaluation purposes but included in the input.

\textbf{Limitations:} This procedure for producing flow labels has two drawbacks. First is that all the points inside a bounding box may not be moving rigidly. While it is important to be aware of this we believe the errors introduced are small since most objects are rigid vehicles and the sampling rate of 10Hz prevents large deviations from the model. Second, there can exist objects that are not captured by the tracking labels but are nonetheless dynamic. This second limitation is the main reason for our interest in self- and un-supervised methods as existing work has demonstrated degradation of supervised performance on objects not included in the training labels~\cite{jund2021scalable}. However, it is supervised training that creates the correlation between performance on an object class and its presence in the labels. For self-supervised methods we expect performance on tracked objects to correlate with those that may be missing.

\section{Local Dynamic Segmentation Model Details}
\begin{figure*}
    \centering
    \includegraphics[width=0.9\textwidth, trim=0cm 7cm 0cm 0,clip]{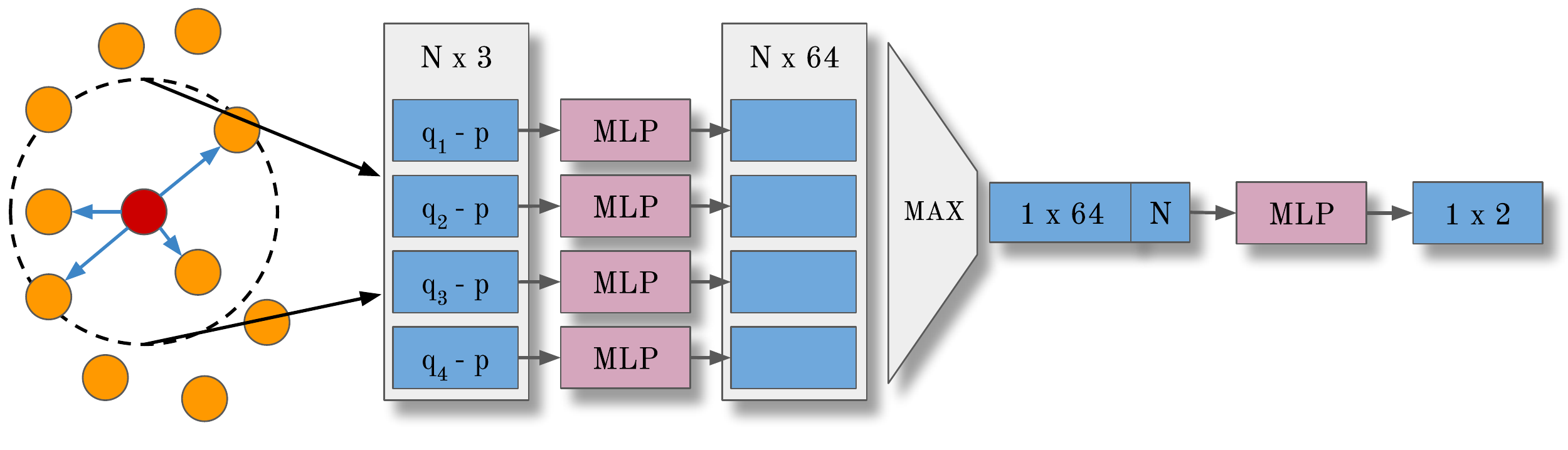}
    \caption{Our local segmentation model only uses information about the relative position of points in a local neighborhood around the predicted point. The local and global MLPs each have two hidden layers of 64 units.}
    \label{fig:kitti-arch}
\end{figure*}

In Sec. 3 of the main paper, we use a local classification model to demonstrate how the sampling pattern of dynamic objects in stereoKITTI makes it trivial to identify moving vs. static objects. Here we give details of the architecture of that model. The architecture is shown in \cref{fig:kitti-arch} and is essentially a very simple version of PointNet\cite{qi2017pointnet} with the transform modules removed. The key component of this model is that the input contains only the relative positions of the query point's neighbors, so no global information is available to the network. 
\section{Optimization Details}
\label{sec:details}
Our baseline uses 6 parameters, which we detail here. In general, the same parameters are used across all datasets (Arogverse, Waymo, NuScenes, lidarKITTI). However, due to the large variations in sparsity in both Waymo and NuScenes two parameters were adjusted: the early stopping criterion, and the DBSCAN epsilon parameter. These adjustments were not found through a parameter search, but by visually inspecting the flow results and clusters respectively.

\textbf{Neural Prior Parameters:} The forwards and backward flow networks have the same structure as \cite{li2021neural} and we optimize them with the Adam\cite{kingma2014adam} optimizer using a learning rate of 0.004 and no weight decay. We stop the optimization when no progress is made for 100 iterations (200 when optimizing on Waymo).

\textbf{RANSAC Parameters:} For all clusters, we do 250 RANSAC iterations and we use an inlier threshold of 0.2. When thresholding on the translation component, we use the same threshold as the dataset's dynamic threshold: \SI[per-mode = symbol]{0.5}{\meter \per \second}.

\textbf{DBSCAN Parameters:} For Argoverse, Waymo and lidarKITTI we use an epsilon parameter of 0.4 but for NuScenes we increase this to 0.8 due to the sparsity. For all datasets, we use a minimum point threshold of 10.

\section{Ablation}
To test the components of our method, we performed an ablation study on Argoverse. The results are shown in \cref{tab:my_label} and show that both Motion Compensation and Rigid Refinement play key roles in improving the performance of the backbone scene flow method. Motion Compensation has the largest impact on estimating dynamic motion since it allows the network to simply assign zero to all background points. The rigid refinement step improves dynamic motion estimates as well, but also has a large impact on static points. This is due to it fixing phantom motion estimates on walls caused by ``swimming'' artifacts.

\begin{table}
    \setlength{\tabcolsep}{2pt}
  \scriptsize
  \centering
  \begin{tabular}{lcccccc}
    \toprule
     & \multicolumn{4}{c}{EPE} & AccR & AccS\\
    \cmidrule(lr){2-5}
    & Avg & Dynamic & \multicolumn{2}{c}{Static} & Dynamic & Dynamic\\
    \cmidrule(lr){4-5}
    & & FG & FG & BG & FG & FG\\
    \midrule
    Backbone & 0.088 & 0.193 & 0.033 & 0.039  & 0.542 & 0.327\\
    w/ Motion Compensation & 0.066 & 0.112 & 0.042 & 0.046 & 0.756 &0.515\\
    w/ Motion \& w/ Rigid Refinement & 0.055 & 0.105 & 0.033 & 0.028 & 0.777 & 0.537\\\bottomrule
    \end{tabular}
    \caption{Ablation of the two main components of our model, motion compensation and rigid refinement.}
    \label{tab:my_label}
\end{table}
\end{document}